\documentclass[11pt,a4paper]{article}
\usepackage{times,latexsym}
\usepackage{url}
\usepackage[T1]{fontenc}
\usepackage{booktabs}
\usepackage{ulem}

\usepackage[acceptedWithA]{tacl2021v1}

\usepackage{amsmath,amsfonts,bm}

\def\eqref#1{equation~\ref{#1}}

\def\1{\bm{1}}

\DeclareMathAlphabet{\mathsfit}{\encodingdefault}{\sfdefault}{m}{sl}
\SetMathAlphabet{\mathsfit}{bold}{\encodingdefault}{\sfdefault}{bx}{n}

\usepackage{hyperref}
\usepackage{url}
\usepackage{tcolorbox}

\usepackage{multirow}
\usepackage{booktabs}

\usepackage{ulem} 

\usepackage{rotating} 
\usepackage{array}    
\usepackage{enumitem}
\usepackage{listings}
\usepackage{pifont}
\PassOptionsToPackage{table,dvipsnames}{xcolor}
\definecolor{DeepGreen}{rgb}{0.0, 0.5, 0.0}

\usepackage{etoolbox}
\makeatletter
\patchcmd{\@citex}{\hskip\@cite@space}{\penalty\@m \hskip\@cite@space}{}{}
\makeatother
\usepackage{colortbl} 
\usepackage{amsmath,amssymb}

\newcommand{\xmark}{\ding{55}}
\newcommand{\cmark}{\ding{51}}

\usepackage{xspace,mfirstuc,tabulary}

\newif\iftaclinstructions
\taclinstructionsfalse 
\iftaclinstructions
\renewcommand{\confidential}{}
\renewcommand{\anonsubtext}{(No author info supplied here, for consistency with
TACL-submission anonymization requirements)}
\newcommand{\instr}
\fi

\iftaclpubformat 

\else

\fi

\definecolor{grey}{gray}{0.5}

\title{RepreGuard: Detecting LLM-Generated Text by Revealing Hidden Representation Patterns}

\author{Xin Chen$^{1,2}$\Thanks{Equal contribution.} ~~~ Junchao Wu$^{1,*}$ ~~~ Shu Yang$^{3}$ ~~~ Runzhe Zhan$^{1}$ ~~~ Zeyu Wu$^{1}$ ~~~ Ziyang Luo$^{4}$ ~~~ \\ \bf Di Wang$^{3}$ ~~~ \bf Min Yang$^{2}$ ~~~ \bf Lidia S. Chao$^{1}$ ~~~ Derek F. Wong$^{1,}$\Thanks{Corresponding author.} \\
$^{1}$NLP$^2$CT Lab, Department of Computer and Information Science, University of Macau\\
$^{2}$Shenzhen Institutes of Advanced Technology, Chinese Academy of Sciences\\
$^{3}$Provable Responsible AI and Data Analytics Lab, KAUST ~~~
$^{4}$Hong Kong Baptist University\\
\texttt{\small nlp2ct.\{xinchen,junchao,runzhe,zeyu\}@gmail.com, \{shu.yang, di.wang\}@kaust.edu.sa} \\
\texttt{\small min.yang@siat.ac.cn, cszyluo@comp.hkbu.edu.hk, \{derekfw,lidiasc\}@um.edu.mo} \\
}

\date{}

\begin{document}
\maketitle

\begin{abstract}

Detecting content generated by large language models (LLMs) is crucial for preventing misuse and building trustworthy AI systems. Although existing detection methods perform well, their robustness in out-of-distribution (OOD) scenarios is still lacking. In this paper, we hypothesize that, compared to features used by existing detection methods, the internal representations of LLMs contain more comprehensive and raw features that can more effectively capture and distinguish the statistical pattern differences between LLM-generated texts (LGT) and human-written texts (HWT). We validated this hypothesis across different LLMs and observed significant differences in neural activation patterns when processing these two types of texts. Based on this, we propose \textbf{RepreGuard}, an efficient statistics-based detection method. Specifically, we first employ a surrogate model to collect representation of LGT and HWT, and extract the distinct activation feature that can better identify LGT. We can classify the text by calculating the projection score of the text representations along this feature direction and comparing with a precomputed threshold. Experimental results show that RepreGuard outperforms all baselines with average $94.92\%$ AUROC on both in-distribution (ID) and OOD scenarios, while also demonstrating robust resilience to various text sizes and mainstream attacks.\footnote{Data and code are publicly available at: \href{https://github.com/NLP2CT/RepreGuard}{https://github.com/NLP2CT/RepreGuard}}

\end{abstract}

\section{Introduction}

\begin{figure*}[!ht]
    \centering    \includegraphics[width=\textwidth, trim=0 0 0 0]{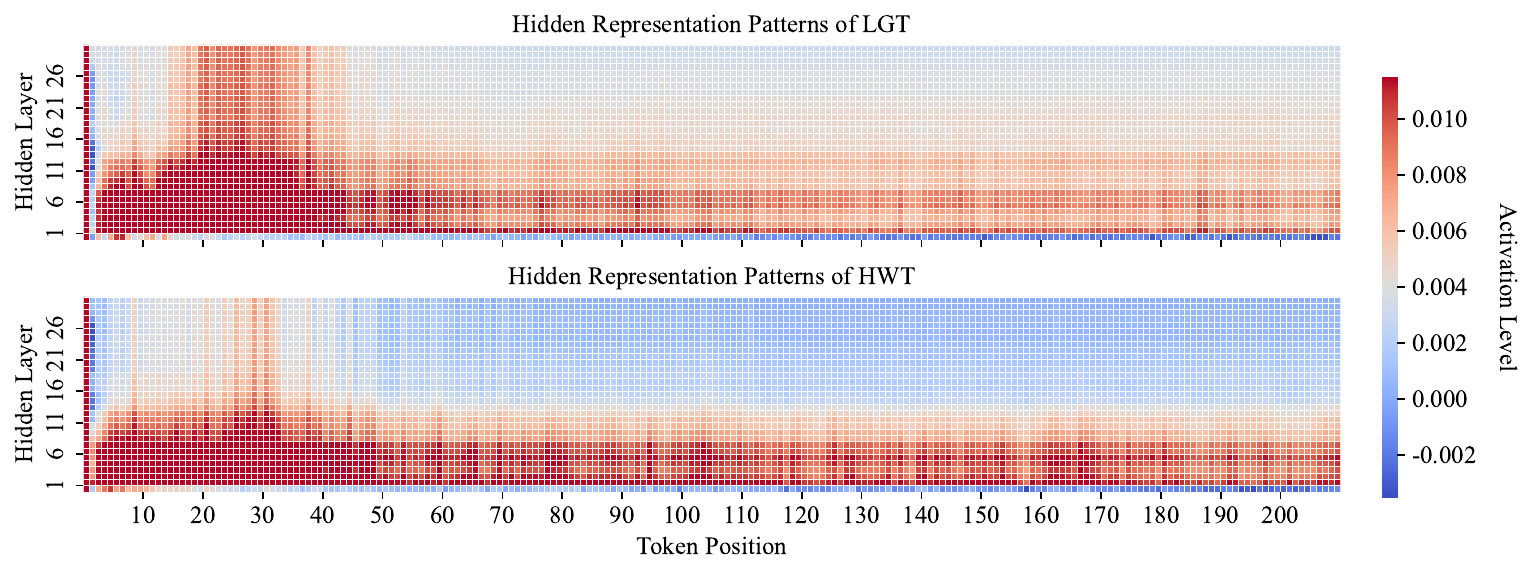}
    \caption{Comparison of Average Hidden Representation Distribution in Llama-3.1-8B Using 1000 Pairs of LGT and HWT from 4 Different LLMs. The \textcolor{red}{red} represents active representations and \textcolor{blue}{blue} represents relatively inactive representations. The depth of the color represents the level of representation activation.}
    \label{fig:Neural_Activity}
\end{figure*}

LLMs demonstrate impressive language understanding and generation capabilities~\cite{blog2022chatgpt,blog2024claude_instant,blog2023palm,blog2024llama3}, enabling them to produce creative and persuasive content that aligns with human preferences~\cite{DBLP:journals/corr/abs-2310-14724}. These capabilities have raised concerns regarding future data regulation, particularly due to biases~\cite{DBLP:journals/corr/abs-2311-04076} and hallucinations~\cite{DBLP:journals/csur/JiLFYSXIBMF23} in LGT. Moreover, the potential misuse of LLMs, such as generating fake news~\cite{DBLP:conf/coling/PagnoniGT22} or facilitating academic dishonesty~\cite{cotton2024chatting}, poses significant risks and challenges to society. To defend against these usage cases, several algorithms have been developed to detect LGT. These primarily include fine-tuning-based classifiers, which involve training an model with extensive labeled data for classification, such as the OpenAI detector~\cite{DBLP:journals/corr/abs-1908-09203}, and statistics-based methods, which identify LGT using feature metrics from specific distributions in a small training dataset, such as DetectGPT~\cite{DBLP:conf/icml/Mitchell0KMF23}.

Fine-tuning-based classifiers generally offer higher accuracy than statistics-based detectors, but require large amounts of labeled data and often struggle to generalize across different generators, making updates costly for new models~\cite{DBLP:journals/corr/abs-2301-07597}. In contrast, statistics-based methods provide better interpretability and only require setting a threshold based on observed distribution patterns in a small sample size, offering stonger reliability for real-world applications~\cite{DBLP:journals/corr/abs-2310-14724}. However, current statistics-based methods perform poorly in both ID and OOD scenarios due to the insufficient robustness in classification feature metrics. For example, varying prompts could control the perplexity of generated text, rendering thresholds from training samples ineffective~\cite{DBLP:conf/icml/HansSCKSGGG24}. This limitation is exacerbated in OOD scenarios, posing significant challenges to the usability of statistics-based detectors, with the growing number of new LLMs.

In response to this challenge, we propose a detection method based on hidden representation to identify texts generated by LLMs. This detection method is motivated by the following hypothesis:  LLMs exhibit distinct hidden representation patterns when processing LGT compared to HWT, due to their different perceptions of statistical patterns in these text types. 

These hidden representation patterns differences can be more explicitly observed in the model's representations, which are higher-dimensional features. This is also where the detection abilities of existing statistics-based detectors come into play, utilizing classification feature metrics such as likelihood, rank, and other variants. Thus, the model's representations may encompass more comprehensive and raw features that can enhance the ability to distinguish between text types, with a stronger potential for identifying LGT. To achieve this, we first employ a surrogate model as an ``observer'' to obtain the representations when processing texts generated by LLMs and those written by humans. We observed that the hidden representation patterns of the two types of texts show distinct differences, which serve as a strong signal for detecting LGT. Then, to filter out noisy features that might hinder LGT detection, we perform dimensionality reduction and modeling on both types of text representations to identify features that maximally retain the distinguishing information. By calculating the projection score of the representation of a given text along this feature direction, which we termed ``RepreScore'', and comparing it to the optimal classification threshold statistically derived, we can determine whether the text was generated by an LLM or written by a human. We call this detection method \textbf{RepreGuard}, which is an efficient detection method that combines the strengths of both statistics-based and fine-tuning-based approaches. RepreGuard exhibits zero-shot characteristics, requiring only a small number of training samples from one LLM source to generalize across texts generated by various types of LLMs.

Experiments on different setups show that RepreGuard outperforms the SOTA RoBERTa-based classifier and the statistics-based method Binoculars in both ID and OOD scenarios, with an average of $11.05\%$ and $05.88\%$ AUROC higher than RoBERTa-based classifier and Binoculars, respectively, and achieve better performance with fewer training samples. In addition, our method demonstrates strong robustness to the generalization on Domains, texts with varied sizes, mainstream attacks including text paraphrasing and perturbation attacks and various sampling methods. It also achieves an excellent balance between effectiveness and resource efficiency.

\section{Related Works}

\paragraph{Statistics-based Detection Methods}

The statistics-based detection method detects LGT by examining the distribution difference of the specified feature metrics between LGT and HWT. This classification is achieved by extracting these feature metrics from the given text and comparing them to thresholds derived from statistics on training datasets, without the need to fine-tune a neural model as classifier. 
Early statistics-based methods focused on calculating feature metrics derived from the model output logits, such as Entropy, Log-Likelihood, and Log-Rank~\cite{DBLP:journals/corr/abs-1908-09203}. Subsequently, \citet{DBLP:conf/emnlp/SuZ0N23} introduced the Log-Likelihood Log-Rank Ratio (LRR), offering a more comprehensive evaluation by calculating the ratio of Log-Likelihood to Log-Rank. Recently, perturbation-based methods have gained significant attention. \citet{DBLP:conf/icml/Mitchell0KMF23} and \citet{DBLP:conf/emnlp/SuZ0N23} used Log-Likelihood and Log-Rank curvature, respectively, to identify LGT, based on the hypothesis that LGT maintains higher Log-Likelihood and Log-Rank after semantic perturbation compared to HWT. \citet{DBLP:conf/iclr/BaoZTY024} enhanced this by replacing the perturbation step in DetectGPT with a more efficient sampling procedure, reducing the computational cost. 
Other methods, like DNA-GPT~\cite{DBLP:conf/iclr/Yang0WPWC24}, use an iterative process where an LLM extends truncated text and assesses authorship via probability differences between the original and extended text. In contrast, GECScore~\cite{wu2025wrote} relies on the finding that LLMs are more likely to correct grammatical errors in human-written text and distinguishes text sources by measuring changes in similarity before and after grammatical error correction.
Binoculars~\cite{DBLP:conf/icml/HansSCKSGGG24} is a novel method that employs a pair of LLMs to calculate the ratio of perplexity and cross-perplexity, measuring how one model’s prediction of the next token surprises another one.

\paragraph{Fine-Tuning-Based Detection Methods}
Another approach is to fine-tune a neural model as a classifier for distinguishing between HWT and LGT, typically requiring a large amount of labeled data. Early efforts focused on fine-tuning pre-trained classifiers for detecting news articles~\cite{DBLP:conf/nips/ZellersHRBFRC19} and social media content \cite{DBLP:journals/corr/abs-2008-00036}. Recent studies~\cite{DBLP:journals/corr/abs-2301-07597,DBLP:journals/corr/abs-2304-07666,DBLP:journals/corr/abs-2305-07969,DBLP:journals/corr/abs-2306-07401} further confirmed the strong performance of fine-tuned language model in identifying LGT. OpenAI's detector, for example, is a fine-tuned RoBERTa-based classifier to perform this task~\cite{DBLP:journals/corr/abs-1908-09203}. 
However, fine-tuning-based classifiers tend to overfit to their training data or the training distribution of the source model, resulting in performance drops when encountering new LLMs or domains data~\cite{DBLP:conf/clef/SarvazyanGRF23}.

\section{RepreGuard}

\begin{figure*}[!ht]
    \centering
    \includegraphics[width=\textwidth, trim=0 0 0 0]{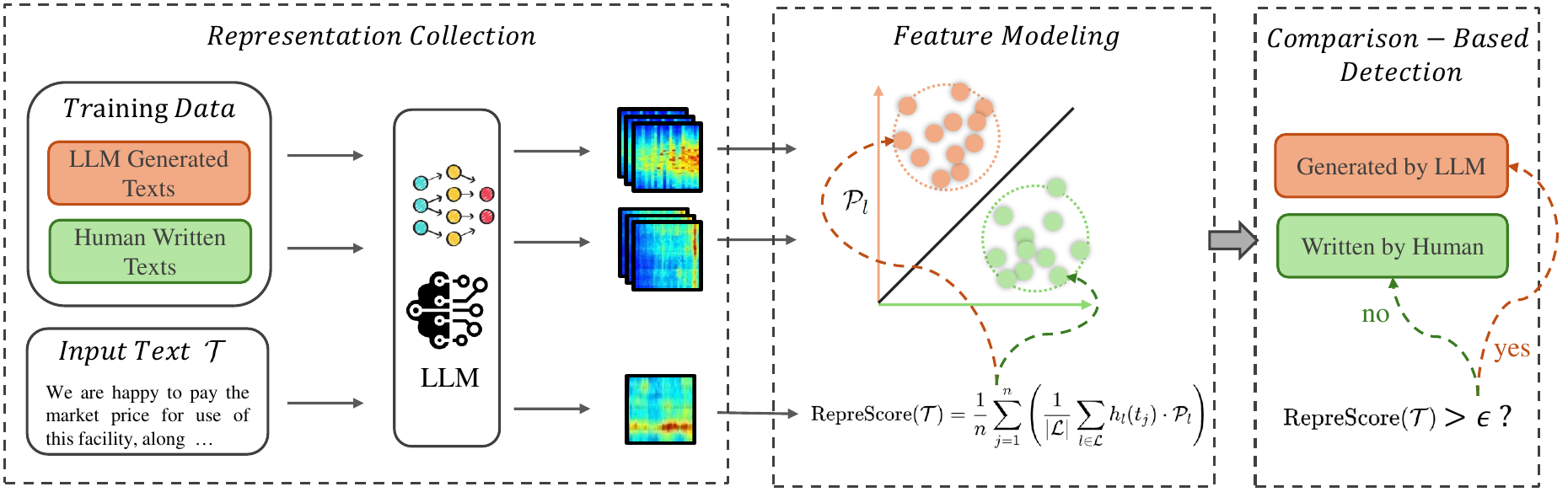}
    \caption{\textbf{The RepreGuard Framework Overview.} The framework processes text through a surrogate model to capture hidden representation patterns and collect representation of LGT and HWT. It employs Principal Component Analysis model distinct activation features, filtering out noise and identifying key features that distinguish LGT. Next, the framework calculates an overall projection score, which we called ``RepreScore'', for the text to quantify how closely a text's activation pattern aligns with LGT representation features. A threshold is then set based on the RepreScores from the training data. In the application tion, if the given text's RepreScore exceeds this threshold, it is more likely to be generated by LLMs.}
    \label{fig:method}
\end{figure*}

\subsection{Preliminary}

\paragraph{Hypothesis} \textit{The premise of RepreGuard is that LLMs perceive the statistical patterns of LGT and HWT differently. Their hidden representation patterns exhibit noticeable differences when observing and processing these two types of text}. 

This hypothesis arises from the idea that there are behavioral differences in the writing processes between LGT and HWT, which can be captured through internal hidden representations, as the internal hidden representations of LLMs typically contain more information and raw features compared to external behaviours. This information can reveal the model's intrinsic understanding of the differences between the LGT and HWT.

\paragraph{Internal Representation of LLMs} 
Recent research~\cite{DBLP:conf/acl/VoitaFN24,DBLP:conf/emnlp/DurraniSDB20,DBLP:journals/corr/abs-2403-11621} suggest that the internal representation mechanisms of LLMs may capture more information. For instance, LLM-Check~\cite{NEURIPS2024_3c1e1fdf} extracts the hidden representations from each layer during response generation, and calculates their covariance matrices (Hidden Score) as an indicator for hallucination detection. \citet{zhang2025do} indirectly reveals the spatial reasoning capabilities encoded in the hidden representations of vision-language models by systematically analyzing output probabilities under continuous variations in input space. \citet{DBLP:conf/acl/TangLH0WZWW24} proposed a method of extracting latent representations as specific concept directions, thereby enabling the effective guidance and control of LLMs towards the concept direction (eg, safety and honesty).

Therefore, RepreGuard is designed to capture the underlying hidden representations that characterise these processes based on the distinct behavioral processes in human writing and AI writing. These distinctions likely result from the statistical patterns internalized by LLMs during training and how these are leveraged in generative tasks. To validate this hypothesis, we follow \citet{DBLP:journals/corr/abs-2310-01405} and compare the neural activation patterns in Llama-3.1-8B when processing 1000 pairs of LGT and HWT from 4 different LLMs, as illustrated in \autoref{fig:Neural_Activity}. Specifically, LGT and HWT exhibit largely similar hidden representations during the early stage of the sequence (approximately tokens 0–20). However, when the number of tokens exceeds 20, their hidden representations begin to diverge substantially, with LGT demonstrating a consistently higher overall activation level compared to HWT. Moreover, at the same sequence positions, LGT and HWT remain relatively similar in the lower layers (approximately layers 1–11), while exhibiting more pronounced differences in layers 11–32.

\subsection{Detecting by Representation Comparisons}

Based on this hypothesis, we propose \textbf{RepreGuard} for detecting LGT using the hidden representation  patterns of LLMs. By analyzing the activation patterns of a surrogate model during the processing of LGT and HWT, we aim to capture subtle but systematic differences in the activation patterns when LLMs process them, and extract the most significant representation feature for the effective detection of LGT. The overview framework of RepreGuard is as illustrated in \autoref{fig:method}.

\paragraph{Representation Collection} 

We first introduce a small training set, formalized as $\{(\mathcal{T}^i_{\text{LGT}}, \mathcal{T}^i_{\text{HWT}}) \mid i \in [1, N]\}$, where $N$ is the number of LGT and HWT pairs in the dataset. We use a surrogate model $\mathcal{M}$ as an ``observer'' to collect the corresponding representation distribution when processing LGT and HWT, to capture the difference in their activation patterns. Specifically, for each text sequence $\mathcal{T} = \{t_1, t_2, ..., t_n\}$ containing $n$ tokens, we use $\mathcal{M}$ with $\mathcal{L}$ layers to ``observe'' it. For each token $t_j$ in $\mathcal{T}$, we collect the neural activation of the model $\mathcal{M}$ at each layer $l$, represented as $h_j^l \in \mathbb{R}^d$, where $d$ is the dimension of the model's hidden state. Based on this, the complete activation of the text $\mathcal{T}$ in model $\mathcal{M}$ can be formalized:
\begin{equation}
\mathcal{A}(\mathcal{T})=\{h_j^l \mid j \in [1, n], l \in [1, \mathcal{L}]\}
\label{eq:activation}
\end{equation}

For each text pair $(\mathcal{T}^i_{\text{LGT}}, \mathcal{T}^i_{\text{HWT}})$, we capture the model's representation activations from the end to the beginning of the token at all layers $\mathcal{L}$. We denote these activations as $\mathcal{A}(\mathcal{T}^i_{\text{LGT}})$, and $\mathcal{A}(\mathcal{T}^i_{\text{HWT}})$.

\paragraph{Feature Modeling}
The difference in activation patterns between LGT and HWT for each text pair can be denoted as $\Delta \mathcal{A}_i$, where:
\begin{equation}
\Delta \mathcal{A}_i = \mathcal{A}(\mathcal{T}^i_{\text{LGT}}) - \mathcal{A}(\mathcal{T}^i_{\text{HWT}})
\end{equation}

For each hidden layer $l$ in $\mathcal{L}$, we capture this difference of neural activations from all text pairs:

\begin{equation}
\Delta \mathcal{A}^l = \{ \Delta \mathcal{A}_i^l \mid i \in [1, N], l \in [1, \mathcal{L}]\ \}
\end{equation}

where $\Delta \mathcal{A}_i^l = h_{\text{LGT}}^{l, i} - h_{\text{HWT}}^{l, i}$ represents the difference for the $i$-th text pair at layer $l$.

To filter out noise and identify key features distinguishing LGT, we perform Principal Component Analysis (PCA) on each $\Delta \mathcal{A}^l$:

\begin{equation}
\mathcal{P}_l = \text{PCA}(\Delta \mathcal{A}^l)
\end{equation}

The resulting $\mathcal{P}_l$ represents the \textit{probing vector} that differentiates between LGT and HWT at layer $l$, demonstrated that different types of text evoke distinct neural activation pattern in the LLM.
We project the activations of each token $t_j$ in text $\mathcal{T}$ onto the probing vector across all layers $\mathcal{L}$, defining this as the \textbf{RepreScore}:
\begin{equation}
\text{RepreScore}(t_j) = \frac{1}{|\mathcal{L}|} \sum_{l \in \mathcal{L}} h_l(t_j) \cdot \mathcal{P}_l
\end{equation}
The overall projection score for the text $\mathcal{T}$ is the mean of its tokens' RepreScores:
\begin{equation}
\text{RepreScore}(\mathcal{T}) = \frac{1}{n} \sum_{j=1}^n \text{RepreScore}(t_j)
\end{equation}

\paragraph{Comparison-Based Detection}

Based on the calculated $\text{RepreScore}(\mathcal{T})$ for each sample in the training dataset, we determine the optimal threshold $\theta$ to balance the true positive rate (TPR) and the false positive rate (FPR):

\begin{equation}
\theta = \arg \max_{\theta'} \left( \text{TPR}(\theta') + (1 - \text{FPR}(\theta')) \right) 
\end{equation}

In the application phase, RepreScore measures how closely the activation pattern of the input text aligns with the unique neural features of LGT identified in the training dataset. For the detection result $\mathcal{S}(\mathcal{T})$, we calculate the $\text{RepreScore}(\mathcal{T})$ and compare it to the optimal threshold $\theta$. If $\text{RepreScore}(\mathcal{T})$ exceeds $\theta$, the input text is more likely to be generated by the LLM:

\begin{equation}
\mathcal{S}(\mathcal{T}) =\left\{
\begin{array}{cl}
\text{LGT}  & \quad \text{if } \text{RepreScore}(\mathcal{T}) > \theta \\
\text{HWT} & \quad \text{otherwise}
\end{array} \right.
\end{equation}

\begin{figure}[!t]
    \centering    \includegraphics[width=0.47\textwidth, trim=0 0 0 0]{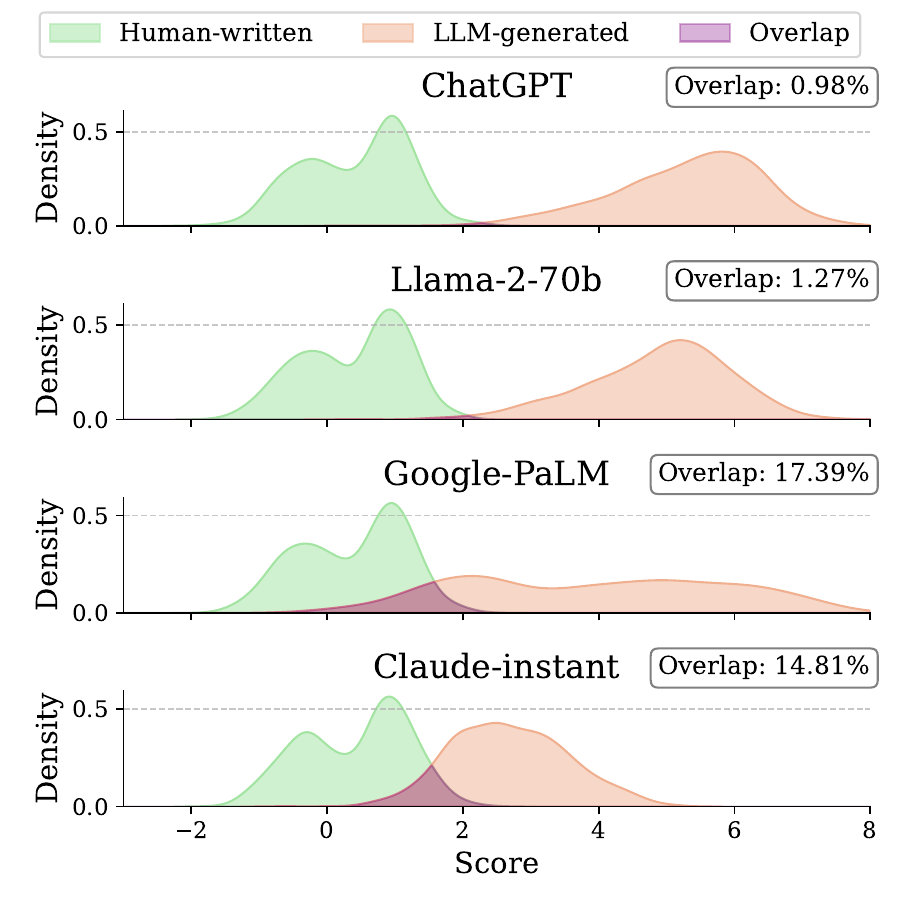}
    \caption{Comparison of RepreScores Distribution for Texts Generated by 4 Different LLMs (1000 Pairs LGT and HWT for Each LLM). The representation features used are modeling on Mutil-LLMs generated texts. The overlap means the overlap probability.}
    \label{fig:RepreScores}
\end{figure}

\paragraph{Effectiveness and Generalization}

Our \textit{RepreGuard} framework effectively detects LGT by extracting features derived from more fundamental and comprehensive representation information. This demonstrates a stronger and more adaptable detection capability. We further validate this capability on LGT generated by four different LLMs and their corresponding HWT. The results in \autoref{fig:RepreScores} clearly show a distinct separation between the RepreScore distributions of LGT and HWT, with a consistent trend across different LLMs. Specifically, the RepreScore for HWT primarily ranges from -2 to 2. Despite variations in text distributions generated by different LLMs, the RepreScore for LGT consistently falls between 0 and 8. This indicates a universal threshold that can be effectively applied across text generated by various types of LLMs, eliminating the need for individual adjustments and highlighting RepreGuard's strong generalization capability.

\section{Experiment}
\subsection{Experiment Setup}
\paragraph{Dataset} 

We utilized the DetectRL benchmark~\cite{DBLP:conf/nips/WuZWY0YC24}, a dataset specifically designed to evaluate the detection capabilities of HWT and LGT in scenarios closely aligned with real-world applications. The dataset comprises data from four domains that are more prone to misuse: academic writing (ArXiv Archive\footnote{\url{https://www.kaggle.com/datasets/spsayakpaul/arxiv-paper-abstracts/data}}), news writing (XSum~\cite{DBLP:conf/emnlp/NarayanCL18}), creative writing (Writing Prompts~\cite{DBLP:conf/acl/LewisDF18}), and social media texts (Yelp Review~\cite{DBLP:conf/nips/ZhangZL15}). For each domain, the dataset includes 2,800 pairs of LGT and HWT samples, with LGT generated using four widely used LLMs: ChatGPT~\cite{blog2022chatgpt}, Claude-instant~\cite{blog2024claude_instant}, Google-PaLM~\cite{blog2023palm}, and Llama-2-70B.\footnote{\url{https://huggingface.co/meta-llama/Llama-2-70b-chat-hf}}

To ensure robust evaluation, we applied the bootstrapping method five times to the dataset of each LLM, sampling a training set consisting of 512 pairs of samples and a test set consisting of 1,000 pairs of samples for each iteration. Additionally, to construct the multi-LLM dataset, we combined the data of four different LLMs and applied the bootstrap method five times equally on the training sets of different LLMs.

\paragraph{Baselines}  

We compare RepreGuard with both fine-tuning-based and statistics-based methods to detect LGT:

\begin{itemize}[label=\textbullet, topsep=5pt, itemsep=3pt, left=2pt]
    \item \textbf{\textit{RoBERTa-based classifier}}~\cite{DBLP:journals/corr/abs-1907-11692}: A supervised approach by fine-tuning pre-training language model (PLM) as a classifier. We use the RoBERTa-base consistent with the OpenAI detector~\cite{DBLP:journals/corr/abs-1908-09203} for training.

    \item \textbf{\textit{LRR}}~\cite{DBLP:conf/emnlp/SuZ0N23}: A statistics-based method based on the ratio of Log-likelihood and Log-rank. We uses GPT-neo-2.7B for scoring following \citet{DBLP:conf/iclr/BaoZTY024} experiments.

    \item \textbf{\textit{DetectGPT}}~\cite{DBLP:conf/icml/Mitchell0KMF23}: A statistics-based zero-shot method based on the Log-probability curvature of the perturbed text. We use T5-small~\cite{DBLP:journals/jmlr/RaffelSRLNMZLL20} to perturb text, and use GPT-Neo-2.7B~\cite{gpt-neo} for scoring following \citet{DBLP:conf/iclr/BaoZTY024}.

    \item \textbf{\textit{Fast-DetectGPT}} (Fast-Detect.;~\citealt{DBLP:conf/iclr/BaoZTY024}): A statistics-based zero-shot method that using a sampling strategy to replace the perturbation strategy of DetectGPT. We use GPT-Neo-2.7B~\cite{gpt-neo} for scoring and GPT-J-6B \cite{gpt-j} for sample generation following the best setting reported.

    \item \textbf{\textit{Straight-forward Detector}} (Str-Detect.): A zero-shot approach that directly asks LLM for both HWT and LGT. This detector is for better comparison with our method using the intrinsic mechanism of LLMs. 

    \item \textbf{\textit{Binoculars}}~\cite{DBLP:conf/icml/HansSCKSGGG24}: A statistics-based zero-shot method that uses a pair of LLMs to calculate the ratio of perplexity and cross-perplexity for detection. We use Falcon-7B and Falcon-7B-Instruct~\cite{DBLP:conf/nips/PenedoMHCACPAL23} for detection following the best setting reported.
\end{itemize}

\paragraph{Metrics}
Following \citet{DBLP:conf/icml/Mitchell0KMF23}, we utilized the AUROC to evaluate the performance of the detector as a binary classification model. In the LGT detection task, the most worrying harm usually comes from false positives, that is, HWT is incorrectly labeled as LGT. Therefore, we also focus on the TPR at low FPR and follow the experimental setting of \citet{DBLP:conf/icml/HansSCKSGGG24} to adopt a standard FPR threshold of 0.01\% TPR (TPR@0.01).

\paragraph{ID and OOD Detection Setting}

Our study examines both the performance of ID and in a strict zero-shot detection scenario, where we use a purely ``OOD'' setting to distinguish LGT from HWT, consistent with the ``out-of-domain'' claims in Binoculars~\cite{DBLP:conf/icml/HansSCKSGGG24} and Ghostbuster~\cite{verma-etal-2024-ghostbuster}. In fact, many previous zero-shot works such as DetectGPT~\cite{DBLP:conf/icml/Mitchell0KMF23} are in different experimental settings for zero-shot detection, where they use the test set to define the threshold and get the best performance on the test set. We argue this experimental setting may limit the development of truly effective statistics-based zero-shot detectors. Therefore, our experimental setting is strictly aligned with the real-world scenario to ensure that the detector is robust and not overly dependent on any particular dataset. We use the training data to make the decision thresholds to detect text generated by unknown LLMs.

\subsection{ID and OOD Performance}

\begin{table*}[!ht]
\vspace{\baselineskip}
\centering
\renewcommand{\arraystretch}{1}
\resizebox{\textwidth}{!}{
\begin{tabular}{ll rr rr rr rr rr rr}
\toprule
\bf Test$\rightarrow$ & \bf Detector$\downarrow$ & \multicolumn{2}{c}{ChatGPT} & \multicolumn{2}{c}{Llama-2-70b} & \multicolumn{2}{c}{Google-PaLM} & \multicolumn{2}{c}{Claude-instant} & \multicolumn{2}{c}{Avg.}\\
\bf Train$\downarrow$ & \bf Metrics$\rightarrow$ & \textit{AUR.} & \textit{TPR.} & \textit{AUR.} & \textit{TPR.} & \textit{AUR.} & \textit{TPR.} & \textit{AUR.} & \textit{TPR.} & \textit{AUR.} & \textit{TPR.} \\
\midrule
\multirow{6}{*}{ChatGPT} 
& Roberta &  \cellcolor{grey!20}$\underline{98.38_{\pm{0.32}}}$ &  \cellcolor{grey!20}$\underline{90.52_{\pm{1.93}}}$ &$81.71_{\pm{2.27}}$ & $54.64_{\pm{16.57}}$ &$74.56_{\pm{0.89}}$ & $20.20_{\pm{12.87}}$ & \cellcolor{grey!20}$\underline{66.74_{\pm{1.49}}}$ &  \cellcolor{grey!20}$\underline{22.36_{\pm{13.26}}}$ &$80.35_{\pm{1.24}}$ & $46.93_{\pm{11.16}}$ \\
& LRR & $92.61_{\pm{0.39}}$ & $26.20_{\pm{0.00}}$ &$95.84_{\pm{0.34}}$ & $86.20_{\pm{0.00}}$ &$81.98_{\pm{0.14}}$ & $39.80_{\pm{0.00}}$ &$57.82_{\pm{0.80}}$ & $0.10_{\pm{0.00}}$ &$82.06_{\pm{0.42}}$ & $38.07_{\pm{0.00}}$ \\
& DetectGPT & $54.87_{\pm{0.25}}$ & $0.11_{\pm{0.14}}$ &$59.21_{\pm{0.53}}$ & $0.66_{\pm{1.77}}$ &$55.29_{\pm{0.37}}$ & $0.08_{\pm{0.06}}$ &$55.92_{\pm{0.61}}$ & $0.00_{\pm{0.00}}$ &$56.32_{\pm{0.44}}$ & $0.21_{\pm{0.49}}$ \\
& Fast-Detect. & $75.65_{\pm{0.28}}$ & $11.60_{\pm{0.00}}$ &$85.49_{\pm{0.31}}$ & $2.50_{\pm{0.00}}$ &$80.36_{\pm{0.63}}$ & $17.70_{\pm{0.00}}$ &$47.29_{\pm{0.44}}$ & $0.00_{\pm{0.00}}$ &$72.20_{\pm{0.42}}$ & $7.95_{\pm{0.00}}$ \\
& Str-Detect. & $52.05_{\pm{0.00}}$ & $0.01_{\pm{0.00}}$ &$52.30_{\pm{0.00}}$ & $0.01_{\pm{0.00}}$ &$56.05_{\pm{0.00}}$ & $0.01_{\pm{0.00}}$ &$57.50_{\pm{0.00}}$ & $0.01_{\pm{0.00}}$ &$54.47_{\pm{0.00}}$ & $0.01_{\pm{0.00}}$ \\
& Binoculars & $97.36_{\pm{0.52}}$ & $40.70_{\pm{0.00}}$ &  \cellcolor{grey!20}$\underline{99.45_{\pm{0.44}}}$ &  \cellcolor{grey!20}$\underline{98.70_{\pm{0.00}}}$ & \cellcolor{blue!20}$\bf{98.03_{\pm{0.34}}}$ & \cellcolor{blue!20}$\bf{89.80_{\pm{0.00}}}$ & $61.86_{\pm{3.64}}$ & $3.40_{\pm{0.00}}$ & \cellcolor{grey!20}$\underline{89.18_{\pm{1.24}}}$ & \cellcolor{grey!20}$\underline{58.15_{\pm{0.00}}}$ \\
& RepreGuard & \cellcolor{blue!20}$\bf{99.84_{\pm{0.12}}}$ &  \cellcolor{blue!20}$\bf{100.00_{\pm{0.00}}}$ &  \cellcolor{blue!20}$\bf{99.55_{\pm{0.12}}}$ &  \cellcolor{blue!20}$\bf{99.26_{\pm{0.11}}}$ &  \cellcolor{grey!20}$\underline{88.67_{\pm{0.65}}}$ &  \cellcolor{grey!20}$\underline{64.66_{\pm{1.26}}}$ &  \cellcolor{blue!20}$\bf{85.00_{\pm{1.40}}}$ &  \cellcolor{blue!20}$\bf{56.12_{\pm{2.20}}}$ &  \cellcolor{blue!20}$\bf{93.26_{\pm{0.57}}}$ &  \cellcolor{blue!20}$\bf{80.01_{\pm{0.89}}}$ \\

\midrule
\multirow{6}{*}{Llama-2-70b}
& Roberta & $95.49_{\pm{0.96}}$ & $71.16_{\pm{4.59}}$ &$94.21_{\pm{2.06}}$ & $63.66_{\pm{10.40}}$ &$86.00_{\pm{2.17}}$ & $43.60_{\pm{2.56}}$ & \cellcolor{grey!20}$\underline{76.98_{\pm{4.04}}}$ & \cellcolor{grey!20}$\underline{22.88_{\pm{5.13}}}$ & \cellcolor{grey!20}$\underline{88.17_{\pm{2.31}}}$ & $50.32_{\pm{5.67}}$ \\
& LRR & $91.88_{\pm{1.05}}$ & $26.20_{\pm{0.00}}$ &$96.47_{\pm{0.52}}$ & $86.20_{\pm{0.00}}$ &$81.65_{\pm{0.55}}$ & $39.80_{\pm{0.00}}$ &$56.20_{\pm{1.77}}$ & $0.10_{\pm{0.00}}$ &$81.55_{\pm{0.97}}$ & $38.07_{\pm{0.00}}$ \\
& DetectGPT & $54.33_{\pm{0.81}}$ & $0.51_{\pm{0.80}}$ &$59.36_{\pm{0.86}}$ & $1.28_{\pm{2.18}}$ &$55.02_{\pm{0.98}}$ & $0.02_{\pm{0.06}}$ &$56.11_{\pm{0.24}}$ & $0.00_{\pm{0.00}}$ &$56.20_{\pm{0.72}}$ & $0.45_{\pm{0.76}}$ \\
& Fast-Detect. & $94.08_{\pm{1.26}}$ & $71.90_{\pm{0.00}}$ &$98.76_{\pm{0.05}}$ & $94.20_{\pm{0.00}}$ &$92.39_{\pm{0.28}}$ & $81.80_{\pm{0.00}}$ &$51.45_{\pm{0.62}}$ & $0.40_{\pm{0.00}}$ &$84.17_{\pm{0.55}}$ & $62.08_{\pm{0.00}}$ \\
& Str-Detect. & $52.05_{\pm{0.00}}$ & $0.01_{\pm{0.00}}$ &$52.30_{\pm{0.00}}$ & $0.01_{\pm{0.00}}$ &$56.05_{\pm{0.00}}$ & $0.01_{\pm{0.00}}$ &$57.50_{\pm{0.00}}$ & $0.01_{\pm{0.00}}$ &$54.47_{\pm{0.00}}$ & $0.01_{\pm{0.00}}$ \\
& Binoculars & $97.94_{\pm{0.74}}$ & \cellcolor{grey!20}$\underline{85.90_{\pm{0.00}}}$ & \cellcolor{blue!20}$\bf{99.63_{\pm{0.06}}}$ & \cellcolor{grey!20}$\underline{98.70_{\pm{0.00}}}$ & \cellcolor{blue!20}$\bf{97.23_{\pm{0.93}}}$ & \cellcolor{blue!20}$\bf{89.80_{\pm{0.00}}}$ & $57.49_{\pm{3.57}}$ & $3.40_{\pm{0.00}}$ & $88.07_{\pm{1.32}}$ & \cellcolor{grey!20}$\underline{69.45_{\pm{0.00}}}$ \\
& RepreGuard &  \cellcolor{blue!20}$\bf{99.54_{\pm{0.08}}}$ &  \cellcolor{blue!20}$\bf{99.38_{\pm{0.14}}}$ & \cellcolor{grey!20}$\underline{99.38_{\pm{0.18}}}$ & \cellcolor{blue!20}$\bf{98.84_{\pm{0.11}}}$ & \cellcolor{grey!20}$\underline{88.84_{\pm{1.28}}}$ & \cellcolor{grey!20}$\underline{77.08_{\pm{0.45}}}$ &  \cellcolor{blue!20}$\bf{84.08_{\pm{3.52}}}$ & \cellcolor{blue!20}$\bf{60.66_{\pm{1.66}}}$ & \cellcolor{blue!20}$\bf{92.96_{\pm{1.27}}}$ & \cellcolor{blue!20}$\bf{83.99_{\pm{0.59}}}$ \\
\midrule
\multirow{6}{*}{Google-PaLM} 
& Roberta & $88.72_{\pm{1.35}}$ & $40.94_{\pm{5.18}}$ &$85.36_{\pm{5.00}}$ & $32.70_{\pm{6.15}}$ &$82.09_{\pm{3.99}}$ & $36.52_{\pm{4.64}}$ & \cellcolor{grey!20}$\underline{72.98_{\pm{4.00}}}$ & \cellcolor{grey!20}$\underline{21.54_{\pm{8.64}}}$ &$82.29_{\pm{3.58}}$ & $32.92_{\pm{6.15}}$ \\
& LRR & $91.40_{\pm{2.01}}$ & $26.20_{\pm{0.00}}$ &$92.84_{\pm{2.78}}$ & $86.20_{\pm{0.00}}$ &$83.13_{\pm{1.53}}$ & $39.80_{\pm{0.00}}$ &$61.11_{\pm{2.05}}$ & $0.10_{\pm{0.00}}$ &$82.12_{\pm{2.09}}$ & $38.07_{\pm{0.00}}$ \\
& DetectGPT & $54.04_{\pm{0.32}}$ & $0.00_{\pm{0.00}}$ &$59.21_{\pm{0.22}}$ & $0.00_{\pm{0.00}}$ &$55.30_{\pm{0.35}}$ & $0.02_{\pm{0.06}}$ &$55.53_{\pm{0.46}}$ & $0.96_{\pm{1.63}}$ &$56.02_{\pm{0.34}}$ & $0.24_{\pm{0.42}}$ \\
& Fast-Detect. & $74.62_{\pm{0.99}}$ & $11.60_{\pm{0.00}}$ &$86.47_{\pm{0.50}}$ & $2.50_{\pm{0.00}}$ &$80.64_{\pm{0.22}}$ & $17.70_{\pm{0.00}}$ &$48.46_{\pm{0.47}}$ & $0.00_{\pm{0.00}}$ &$72.55_{\pm{0.54}}$ & $7.95_{\pm{0.00}}$ \\
& Str-Detect. & $52.05_{\pm{0.00}}$ & $0.01_{\pm{0.00}}$ &$52.30_{\pm{0.00}}$ & $0.01_{\pm{0.00}}$ &$56.05_{\pm{0.00}}$ & $0.01_{\pm{0.00}}$ &$57.50_{\pm{0.00}}$ & $0.01_{\pm{0.00}}$ &$54.47_{\pm{0.00}}$ & $0.01_{\pm{0.00}}$ \\
& Binoculars & \cellcolor{blue!20}$\bf{98.56_{\pm{0.35}}}$ & \cellcolor{grey!20}$\underline{85.90_{\pm{0.00}}}$ & \cellcolor{blue!20}$\bf{99.58_{\pm{0.20}}}$ & \cellcolor{grey!20}$\underline{98.70_{\pm{0.00}}}$ & \cellcolor{blue!20}$\bf{97.98_{\pm{0.38}}}$ & \cellcolor{blue!20}$\bf{89.80_{\pm{0.00}}}$ & $61.15_{\pm{2.49}}$ & $3.40_{\pm{0.00}}$ & \cellcolor{grey!20}$\underline{89.32_{\pm{0.86}}}$ & \cellcolor{grey!20}$\underline{69.45_{\pm{0.00}}}$ \\
& RepreGuard & \cellcolor{grey!20}$\underline{98.39_{\pm{0.31}}}$ & \cellcolor{blue!20}$\bf{99.36_{\pm{0.07}}}$ & \cellcolor{grey!20}$\underline{98.53_{\pm{0.28}}}$ & \cellcolor{blue!20}$\bf{99.16_{\pm{0.07}}}$ & \cellcolor{grey!20}$\underline{93.36_{\pm{0.27}}}$ & \cellcolor{grey!20}$\underline{79.88_{\pm{3.43}}}$ & \cellcolor{blue!20}$\bf{90.57_{\pm{1.06}}}$ & \cellcolor{blue!20}$\bf{56.90_{\pm{2.20}}}$ & \cellcolor{blue!20}$\bf{95.21_{\pm{0.48}}}$ & \cellcolor{blue!20}$\bf{83.82_{\pm{1.44}}}$ \\

\midrule
\multirow{6}{*}{Claude-instant} 
& Roberta & $88.63_{\pm{1.34}}$ & $28.50_{\pm{11.74}}$ &$77.45_{\pm{4.73}}$ & $23.70_{\pm{10.19}}$ &$74.90_{\pm{1.76}}$ & $14.58_{\pm{13.14}}$ &\cellcolor{grey!20}$\underline{88.07_{\pm{3.89}}}$ & \cellcolor{grey!20}$\underline{36.56_{\pm{4.83}}}$ &$82.26_{\pm{2.93}}$ & $25.84_{\pm{9.98}}$ \\
& LRR & $86.33_{\pm{4.23}}$ & $26.20_{\pm{0.00}}$ &$87.46_{\pm{4.43}}$ & $86.20_{\pm{0.00}}$ &$81.19_{\pm{3.34}}$ & $39.80_{\pm{0.00}}$ &$63.74_{\pm{1.07}}$ & $0.10_{\pm{0.00}}$ &$79.68_{\pm{3.27}}$ & $38.07_{\pm{0.00}}$ \\
& DetectGPT & $53.95_{\pm{0.84}}$ & $0.02_{\pm{0.06}}$ &$57.90_{\pm{1.09}}$ & $0.00_{\pm{0.00}}$ &$54.33_{\pm{1.13}}$ & $0.06_{\pm{0.07}}$ &$55.97_{\pm{0.38}}$ & $0.00_{\pm{0.00}}$ &$55.54_{\pm{0.86}}$ & $0.02_{\pm{0.03}}$ \\
& Fast-Detect. & $81.27_{\pm{5.40}}$ & $71.90_{\pm{0.00}}$ &$82.35_{\pm{4.48}}$ & $94.20_{\pm{0.00}}$ &$80.85_{\pm{4.70}}$ & $81.80_{\pm{0.00}}$ &$61.35_{\pm{0.35}}$ & $0.40_{\pm{0.00}}$ &$76.46_{\pm{3.73}}$ & $62.08_{\pm{0.00}}$ \\
& Str-Detect. & $52.05_{\pm{0.00}}$ & $0.01_{\pm{0.00}}$ &$52.30_{\pm{0.00}}$ & $0.01_{\pm{0.00}}$ &$56.05_{\pm{0.00}}$ & $0.01_{\pm{0.00}}$ &$57.50_{\pm{0.00}}$ & $0.01_{\pm{0.00}}$ &$54.47_{\pm{0.00}}$ & $0.01_{\pm{0.00}}$ \\
& Binoculars & \cellcolor{grey!20}$\underline{92.36_{\pm{1.95}}}$ & \cellcolor{grey!20}$\underline{85.90_{\pm{0.00}}}$ & \cellcolor{grey!20}$\underline{93.18_{\pm{1.01}}}$ & \cellcolor{blue!20}$\bf{98.70_{\pm{0.00}}}$ & \cellcolor{blue!20}$\bf{92.83_{\pm{1.18}}}$ & \cellcolor{blue!20}$\bf{89.80_{\pm{0.00}}}$ & $73.92_{\pm{0.17}}$ & $3.40_{\pm{0.00}}$ & \cellcolor{grey!20}$\underline{88.07_{\pm{1.08}}}$ & \cellcolor{grey!20}$\underline{69.45_{\pm{0.00}}}$ \\
& RepreGuard & \cellcolor{blue!20}$\bf{97.20_{\pm{1.39}}}$ & \cellcolor{blue!20}$\bf{96.92_{\pm{0.49}}}$ & \cellcolor{blue!20}$\bf{97.51_{\pm{0.99}}}$ & \cellcolor{grey!20}$\underline{97.16_{\pm{0.40}}}$ & \cellcolor{grey!20}$\underline{87.77_{\pm{2.37}}}$ & \cellcolor{grey!20}$\underline{63.04_{\pm{0.73}}}$ & \cellcolor{blue!20}$\bf{92.76_{\pm{0.49}}}$ & \cellcolor{blue!20}$\bf{56.22_{\pm{5.22}}}$ & \cellcolor{blue!20}$\bf{93.81_{\pm{1.31}}}$ & \cellcolor{blue!20}$\bf{78.34_{\pm{1.71}}}$ \\

\midrule
\multirow{6}{*}{Multi-LLMs} 
& Roberta & $92.13_{\pm{2.80}}$ & $60.42_{\pm{10.23}}$ &$87.84_{\pm{4.06}}$ & $45.90_{\pm{15.30}}$ &$82.07_{\pm{3.28}}$ & $31.16_{\pm{5.57}}$ &\cellcolor{grey!20}$\underline{82.99_{\pm{5.42}}}$ & \cellcolor{grey!20}$\underline{25.64_{\pm{4.52}}}$ &$86.26_{\pm{3.89}}$ & $40.78_{\pm{8.91}}$ \\
& LRR & $92.78_{\pm{0.27}}$ & $26.20_{\pm{0.00}}$ &$94.98_{\pm{0.27}}$ & $86.20_{\pm{0.00}}$ &$81.94_{\pm{0.05}}$ & $39.80_{\pm{0.00}}$ &$59.93_{\pm{0.41}}$ & $0.10_{\pm{0.00}}$ &$82.41_{\pm{0.25}}$ & $38.07_{\pm{0.00}}$ \\
& DetectGPT & $54.92_{\pm{0.13}}$ & $0.11_{\pm{0.14}}$ &$59.05_{\pm{0.22}}$ & $0.02_{\pm{0.06}}$ &$55.29_{\pm{0.37}}$ & $0.08_{\pm{0.06}}$ &$55.98_{\pm{0.35}}$ & $0.00_{\pm{0.00}}$ &$56.31_{\pm{0.27}}$ & $0.05_{\pm{0.06}}$ \\
& Fast-Detect. & $94.98_{\pm{1.72}}$ & $71.90_{\pm{0.00}}$ &$95.97_{\pm{2.65}}$ & $94.20_{\pm{0.00}}$ &$93.19_{\pm{1.48}}$ & $81.80_{\pm{0.00}}$ &$56.57_{\pm{3.33}}$ & $0.40_{\pm{0.00}}$ &$85.18_{\pm{2.30}}$ & $62.08_{\pm{0.00}}$ \\
& Str-Detect. & $52.05_{\pm{0.00}}$ & $0.01_{\pm{0.00}}$ &$52.30_{\pm{0.00}}$ & $0.01_{\pm{0.00}}$ &$56.05_{\pm{0.00}}$ & $0.01_{\pm{0.00}}$ &$57.50_{\pm{0.00}}$ & $0.01_{\pm{0.00}}$ &$54.47_{\pm{0.00}}$ & $0.01_{\pm{0.00}}$ \\
& Binoculars & \cellcolor{grey!20}$\underline{97.95_{\pm{0.89}}}$ & \cellcolor{grey!20}$\underline{85.90_{\pm{0.00}}}$ & \cellcolor{grey!20}$\underline{98.05_{\pm{1.17}}}$ & \cellcolor{grey!20}$\underline{98.70_{\pm{0.00}}}$ & \cellcolor{blue!20}$\bf{97.58_{\pm{0.70}}}$ & \cellcolor{blue!20}$\bf{89.80_{\pm{0.00}}}$ & $68.59_{\pm{3.55}}$ & $3.40_{\pm{0.00}}$ & \cellcolor{grey!20}$\underline{90.54_{\pm{1.58}}}$ & \cellcolor{grey!20}$\underline{69.45_{\pm{0.00}}}$ \\
& RepreGuard & \cellcolor{blue!20}$\bf{98.00_{\pm{0.43}}}$ & \cellcolor{blue!20}$\bf{99.26_{\pm{0.11}}}$ & \cellcolor{blue!20}$\bf{98.44_{\pm{0.24}}}$ & \cellcolor{blue!20}$\bf{99.20_{\pm{0.12}}}$ & \cellcolor{grey!20}$\underline{92.35_{\pm{0.46}}}$ & \cellcolor{grey!20}$\underline{74.74_{\pm{4.52}}}$ & \cellcolor{blue!20}$\bf{93.40_{\pm{0.74}}}$ & \cellcolor{blue!20}$\bf{56.52_{\pm{1.04}}}$ & \cellcolor{blue!20}$\bf{95.55_{\pm{0.47}}}$ & \cellcolor{blue!20}$\bf{82.43_{\pm{1.45}}}$ \\

\bottomrule
\end{tabular}
}
\caption{ID and OOD Performance Comparison of Detection Algorithms on Different Train and Eval Settings. RepreGuard shows the strongest detection performance on all settings, with an average of $96.34_{\pm{0.27}}\%$ and $93.49_{\pm{1.13}}\%$ AUROC (\textit{AUR.}), and $83.74_{\pm{1.56}}\%$ and $81.13_{\pm{2.11}}\%$ TPR@0.01 (\textit{TPR.}) in ID and OOD respectively. We conduct 5 rounds of bootstrapping and report the mean, standard deviation, and 95\% confidence interval. Here, the subscript represents the standard deviation (e.g., 99.84 ± 0.12 indicates a mean value of 99.84 with a standard deviation of 0.12). The \colorbox{blue!20}{blue background} or \textbf{bold} indicates the best performance and the \colorbox{grey!20}{grey background} or \underline{underline} indicates the second best.}
\label{tab:main_experiment}
\end{table*}

Our experiments evaluate both ID and OOD performance of the detectors, as shown in \autoref{tab:main_experiment}. The results demonstrate that RepreGuard achieves the best overall performance, excelling in both ID and OOD scenarios, achieving an average of $94.92_{\pm{0.70}}\%$ and $82.44_{\pm{1.84}}\%$ in AUROC and TPR@0.01, respectively.

\paragraph{ID Performance}
ID performance refers to the detectors' ability to detect instances within the same distribution as the training dataset. The results show that RepreGuard outperforms other detectors in both AUROC and TPR@0.01 in ID setting. Specifically, it achieves $96.34_{\pm{0.27}}\%$ AUROC and $83.74_{\pm{1.56}}\%$ TPR@0.01, demonstrating its strength in detecting data from the same distribution as the training datasets. Binoculars is the second-best performing method in most settings, particularly in the AUROC metric, with an average AUROC of 92.16\% while its TPR@0.01 is only achieving an average of 58.15\%. Fine-tuning classifiers based on RoBERTa generally performs well to some extent, especially in terms of AUROC, with values rarely approaching RepreGuard (e.g., $98.38_{\pm{0.32}}\%$ AUROC on ChatGPT generated text), while its overall performance can be unstable and may experience significant drops (e.g., $82.09_{\pm{3.99}}\%$ AUROC on on Google-PaLM generated text), and its average TPR@0.01 is $56.82_{\pm{5.45}}\%$. Additionally, LRR, DetectGPT, and Fast-Detect perform poorly in TPR@0.01 across most datasets, suggesting they struggle in achieving accurate LGT detection within the distribution.

\begin{table*}[ht]
\centering
\resizebox{0.9\textwidth}{!}{
\begin{tabular}{l cc cc cc cc cc cc} 
\toprule
\bf Train$\rightarrow$ & \multicolumn{2}{c}{ChatGPT} & \multicolumn{2}{c}{Llama-2-70b} & \multicolumn{2}{c}{Google-PaLM} & \multicolumn{2}{c}{Claude-instant} & \multicolumn{2}{c}{Mutil-LLMs} & \multicolumn{2}{c}{Avg.} \\
\bf Surrogate Model$\downarrow$ Metrics$\rightarrow$ & \textit{AUR.} & \textit{TPR.} & \textit{AUR.} & \textit{TPR.} & \textit{AUR.} & \textit{TPR.} & \textit{AUR.} & \textit{TPR.} & \textit{AUR.} & \textit{TPR.} & \textit{AUR.} & \textit{TPR.}  \\
\midrule
LLama-3.1-8B & \cellcolor{grey!20}\underline{93.20} & \cellcolor{blue!20}\bf{79.70} & \cellcolor{grey!20}\underline{94.35} & \cellcolor{blue!20}\bf{81.80} & \cellcolor{grey!20}\underline{95.05} & \cellcolor{grey!20}\underline{84.80} & 94.80 & \cellcolor{blue!20}\bf{77.10} & \cellcolor{grey!20}\underline{96.30} & \cellcolor{grey!20}\underline{81.20} & \cellcolor{blue!20}\bf{94.74} & \cellcolor{blue!20}\bf{80.92} \\
LLama-3.1-8B-Instruct & 90.20 & \cellcolor{grey!20}\underline{77.10} & 93.85 & 76.50 & 93.40 & 79.40 & 94.75 & 63.10 & 95.75 & 77.20 & 93.59 & \cellcolor{grey!20}\underline{74.66} \\
Llama-3-8B & 92.45 & 69.70 & 90.70 & 70.30 & 94.35 & 73.50 & 93.70 & 65.70 & 94.95 & 69.70 & 93.23 & 69.78 \\
Llama-3-8B-Instruct & 91.45 & 65.10 & 92.30 & 56.80 & 95.05 & 69.10 & 94.60 & 57.60 & 95.45 & 65.30 & 93.77 & 62.78 \\
Llama-2-7B & 87.85 & 66.20 & 92.30 & \cellcolor{grey!20}\underline{78.70} & \cellcolor{blue!20}\bf{95.40} & \cellcolor{blue!20}\bf{85.20} & 94.45 & 65.70 & \cellcolor{blue!20}\bf{96.85} & \cellcolor{blue!20}\bf{85.20} & 93.37 & 76.20 \\
Llama-2-7B-Instruct & 90.60 & 65.70 & 90.70 & 80.10 & 92.10 & 45.90 & \cellcolor{grey!20}\underline{96.10} & 57.50 & 95.95 & 84.00 & 93.09 & 66.64 \\
Mistral-7B & 85.95 & 61.80 & 85.35 & 46.80 & 78.70 & 27.80 & 88.65 & 63.90 & 85.85 & 47.80 & 84.90 & 49.62 \\
Mistral-7B-Instruct & \cellcolor{blue!20}\bf{93.85} & 55.80 & 92.35 & 40.70 & 89.05 & 32.90 & 95.25 & 69.80 & 93.35 & 45.40 & 92.77 & 48.92 \\
Falcon-7B & 67.55 & 2.30 & 86.50 & 46.00 & 89.70 & 50.50 & 89.80 & 27.90 & 69.15 & 2.10 & 80.54 & 25.76 \\
Falcon-7B-Instruct  & 74.60 & 0.00 & 92.20 & 65.40 & 90.05 & 67.10 & 92.45 & \cellcolor{grey!20}\underline{76.20} & 72.60 & 0.00 & 84.38 & 41.74 \\
Gemma-7B & 78.50 & 2.90 & 78.20 & 0.60 & 77.70 & 0.50 & 75.30 & 0.00 & 77.55 & 0.00 & 77.45 & 0.80 \\
Gemma-7B-Instruct & 86.50 & 69.60 & 90.95 & 68.40 & 78.95 & 18.10 & 74.70 & 3.40 & 81.70 & 32.00 & 82.56 & 38.30 \\
Gemma-2B & 86.55 & 70.20 & 90.80 & 74.60 & 89.80 & 64.80 & 91.40 & 36.90 & 92.40 & 72.50 & 90.19 & 63.80 \\
Gemma-2B-Instruct & 93.35 & 75.30 & \cellcolor{blue!20}\bf{94.70} & 81.10 & 91.35 & 72.10 & 90.00 & 65.80 & 94.35 & 76.70 & 92.75 & 74.20 \\
Phi-3-Mini-4K-Instruct & 85.35 & 36.10 & 89.90 & 23.00 & 89.40 & 2.10 & 89.30 & 1.00 & 91.85 & 3.70 & 89.16 & 13.18 \\
Phi-2 & 92.80 & 75.80 & 93.55 & 69.20 & 94.85 & 68.40 & \cellcolor{blue!20}\bf{96.10} & 54.50 & 93.40 & 58.80 & \cellcolor{grey!20}\underline{94.14} & 65.34 \\
GPT-J-6B & 83.10 & 9.50 & 89.45 & 64.80 & 89.75 & 55.80 & 86.95 & 24.40 & 85.30 & 15.50 & 86.91 & 34.00 \\
GPT-Neo-2.7B & 48.90 & 0.10 & 50.00 & 0.00 & 50.00 & 0.10 & 50.00 & 0.10 & 49.75 & 0.00 & 49.73 & 0.06 \\
\bottomrule
\end{tabular}
}
\caption{Performance Comparison of RepreGuard Using Different Surrogate Models on 1000 ``HWT-LGT'' Pairs from 4 Different LLMs. The \colorbox{blue!20}{blue background} or \textbf{bold} indicates the best performance and the \colorbox{grey!20}{grey background} or \underline{underline} indicates the second best.}
\label{tab:surrogate_model_result}
\end{table*}

\paragraph{OOD Performance}
OOD performance measures how well a model detects unseen data that differs from the training set. RepreGuard demonstrates exceptional performance in OOD scenarios, with significantly smaller drops in both AUROC and TPR@0.01 compared to other detectors. This is particularly evident in the testing on Claude-instant. For instance, when trained on Google-PaLM and tested on Claude-instant, RepreGuard achieves an AUROC of $90.57_{\pm{1.06}}\%$ and a TPR@0.01 of $56.22_{\pm{5.22}}\%$. In contrast, Binoculars experiences a drop to an AUROC of $61.15_{\pm{2.49}}\%$ and a TPR@0.01 of $3.40_{\pm{0.00}}\%$, while the RoBERTa classifier drops to an AUROC of $72.98_{\pm{4.00}}\%$ and a TPR@0.01 of $21.56_{\pm{8.64}}\%$. Notably, Binoculars performs particularly well on the Google-PaLM test set, highlighting its preference for certain models. Furthermore, although the RoBERTa-based classifier demonstrates relatively high AUROC performance in certain scenarios, such as achieving $95.49_{\pm{0.96}}\%$ when trained on Llama-2-70b and tested on ChatGPT, its overall performance appears relatively unstable. The combined AUROC and TPR@0.01 are only $81.27_{\pm{1.33}}\%$ and $41.36_{\pm{3.87}}\%$ respectively, highlighting the instability of its performance across different scenarios.

\subsection{Ablation Study}

\paragraph{Surrogate Model}

We evaluated the performance of RepreGuard in detection using surrogate models of different sizes and structures on 1000 random sampled “HWT-LGT” pairs from 4 different LLMs. The results in \autoref{tab:surrogate_model_result} shown LLama-3.1-8B achieved the best performance with 94.74\% AUROC and 80.92\% TPR@0.01, and was chosen as the example surrogate model in our paper. We find that LLMs with large sizes (7B or above), such as Llama-2-7B and Mistral-7B, consistently performed well. However, smaller models do not necessarily mean poor performance. For instance, phi-2 (2.7B) and Gemma-2B-Instruct outperformed most 7B models, achieving an AUROC of 94.14\% and 92.75\%, suggesting that our approach can be effectively deployed even with limited computational resources. Additionally, instruction-tuned models generally performed better than their non-instruction version, though this is not always the case (e.g., LLama-3.1-8B). Therefore, the performance of surrogate models may relate to their architecture and training data, requiring proper evaluation to determine model effectiveness.


\begin{figure}[!ht]
    \centering
    \includegraphics[width=0.48\textwidth, trim=0 0 0 0]{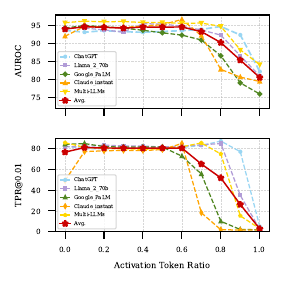}
    \caption{Effect of Activation Token Ratio in Terms of AUROC and TPR@0.01 (\%) by 4 Different LLMs (250 ``HWT-LGT'' Pairs for Each LLM). The model name in the legend refers to the model-generated text used for representation features modeling and threshold setting.}
    \label{fig:activation_representation_ratio}
\end{figure}

\paragraph{Activation Token Ratio}

\begin{table*}[!ht]
\vspace{\baselineskip}
\centering
\resizebox{0.95\textwidth}{!}{
\begin{tabular}{ll cc cc cc cc cc cc cc}
\toprule
\bf Shots$\rightarrow$ & \bf Detector$\downarrow$ & \multicolumn{2}{c}{16} & \multicolumn{2}{c}{32} & \multicolumn{2}{c}{64} & \multicolumn{2}{c}{128} & \multicolumn{2}{c}{256} & \multicolumn{2}{c}{512} & \multicolumn{2}{c}{1024} \\

\bf Train$\downarrow$ & \bf Metrics$\rightarrow$ & \textit{AUR.} & \textit{TPR.} & \textit{AUR.} & \textit{TPR.} & \textit{AUR.} & \textit{TPR.} & \textit{AUR.} & \textit{TPR.} & \textit{AUR.} & \textit{TPR.} & \textit{AUR.} & \textit{TPR.} & \textit{AUR.} & \textit{TPR.} \\
\midrule
\multirow{6}{*}{ChatGPT} 
& Roberta & 60.55 & 0.40 & 42.05 & 0.00 & 64.10 & 0.00 & 68.05 & 0.90 & 84.30 & 5.90 & 86.40 & 48.00 & 88.55 & 44.70 \\
& LRR & 80.55 & 31.50 & 81.10 & 31.50 & 82.30 & 31.50 & 82.85 & 31.50 & 83.35 & 31.50 & 83.45 & 31.50 & 82.75 & 31.50 \\
& DetectGPT & 53.68 & 0.15 & 54.07 & 0.04 & 54.02 & 0.19 & 55.89 & 0.08 & 56.22 & 0.13 & 57.88 & 0.06 & 57.78 & 0.17 \\
& Fast-Detect. & \cellcolor{blue!20}\bf{84.80} & 58.80 & \cellcolor{blue!20}\bf{84.85} & 58.80 & 84.75 & 58.80 & 84.70 & 58.80 & 84.75 & 58.80 & 84.70 & 58.80 & 84.70 & 58.80 \\
& Binoculars & \cellcolor{grey!20}\underline{84.70} & \cellcolor{grey!20}\underline{72.60} & \cellcolor{grey!20}\underline{84.70} & \cellcolor{grey!20}\underline{72.60} & \cellcolor{grey!20}\underline{88.00} & \cellcolor{grey!20}\underline{72.60} & \cellcolor{grey!20}\underline{88.00} & \cellcolor{grey!20}\underline{72.60} & \cellcolor{grey!20}\underline{90.25} & \cellcolor{grey!20}\underline{72.60} & \cellcolor{grey!20}\underline{88.45} & \cellcolor{grey!20}\underline{72.60} & \cellcolor{grey!20}\underline{90.15} & \cellcolor{grey!20}\underline{72.60} \\
& RepreGuard & 83.20 & \cellcolor{blue!20}\bf{81.00} & 82.55 & \cellcolor{blue!20}\bf{83.70} & \cellcolor{blue!20}\bf{91.05} & \cellcolor{blue!20}\bf{81.60} & \cellcolor{blue!20}\bf{90.50} & \cellcolor{blue!20}\bf{81.00} & \cellcolor{blue!20}\bf{93.15} & \cellcolor{blue!20}\bf{78.60} & \cellcolor{blue!20}\bf{93.20} & \cellcolor{blue!20}\bf{79.70} & \cellcolor{blue!20}\bf{93.85} & \cellcolor{blue!20}\bf{79.40} \\
\midrule 
\multirow{6}{*}{Llama-2-70b} 
& Roberta & 52.55 & 0.20 & 52.15 & 0.20 & 41.65 & 0.00 & 60.45 & 0.00 & 83.85 & 27.20 & 85.85 & 30.30 & \cellcolor{blue!20}\bf{95.50} & \cellcolor{grey!20}\underline{73.40} \\
& LRR & 75.40 & 31.50 & 81.10 & 31.50 & 82.80 & 31.50 & 82.95 & 31.50 & 82.45 & 31.50 & 82.45 & 31.50 & 82.80 & 31.50 \\
& DetectGPT & 54.12 & 0.11 & 53.89 & 0.07 & 54.33 & 0.18 & 56.01 & 0.02 & 55.76 & 0.16 & 57.23 & 0.09 & 58.09 & 0.13 \\
& Fast-Detect. & 79.35 & 58.80 & \cellcolor{grey!20}\underline{84.10} & 58.80 & 84.20 & 58.80 & 84.50 & 58.80 & 85.00 & 58.80 & 84.70 & 58.80 & 84.70 & 58.80 \\
& Binoculars & \cellcolor{grey!20}\underline{83.00} & \cellcolor{grey!20}\underline{72.60} & 83.00 & \cellcolor{grey!20}\underline{72.60} & \cellcolor{grey!20}\underline{86.10} & \cellcolor{grey!20}\underline{72.60} & \cellcolor{grey!20}\underline{86.10} & \cellcolor{grey!20}\underline{72.60} & \cellcolor{grey!20}\underline{87.25} & \cellcolor{grey!20}\underline{72.60} & \cellcolor{grey!20}\underline{87.25} & \cellcolor{grey!20}\underline{72.60} & 87.25 & 72.60 \\
& RepreGuard & \cellcolor{blue!20}\bf{89.55} & \cellcolor{blue!20}\bf{79.20} & \cellcolor{blue!20}\bf{89.55} & \cellcolor{blue!20}\bf{78.70} & \cellcolor{blue!20}\bf{96.35} & \cellcolor{blue!20}\bf{77.90} & \cellcolor{blue!20}\bf{91.85} & \cellcolor{blue!20}\bf{82.00} & \cellcolor{blue!20}\bf{95.50} & \cellcolor{blue!20}\bf{81.70} & \cellcolor{blue!20}\bf{94.35} & \cellcolor{blue!20}\bf{81.80} & \cellcolor{grey!20}\underline{94.05} & \cellcolor{blue!20}\bf{82.70} \\
\midrule
\multirow{6}{*}{Google-PaLM} 
& Roberta & 55.45 & 0.50 & 61.10 & 0.10 & 66.30 & 0.00 & 71.25 & 2.40 & 77.50 & 2.80 & \cellcolor{grey!20}\underline{89.40} & 22.90 & \cellcolor{blue!20}\bf{95.45} & 64.50 \\
& LRR & 78.95 & 31.50 & 81.15 & 31.50 & 81.35 & 31.50 & 82.15 & 31.50 & 82.30 & 31.50 & 83.35 & 31.50 & 83.20 & 31.50 \\
& DetectGPT & 54.12 & 0.08 & 54.45 & 0.19 & 54.89 & 0.12 & 55.76 & 0.04 & 56.34 & 0.17 & 57.21 & 0.06 & 57.95 & 0.15 \\
& Fast-Detect. & 76.20 & 58.80 & 83.50 & 58.80 & 85.35 & 58.80 & 85.35 & 58.80 & 85.25 & 58.80 & 85.25 & 58.80 & 85.30 & 58.80 \\
& Binoculars & \cellcolor{grey!20}\underline{86.00} & \cellcolor{grey!20}\underline{72.60} & \cellcolor{grey!20}\underline{88.95} & \cellcolor{grey!20}\underline{72.60} & \cellcolor{grey!20}\underline{88.95} & \cellcolor{grey!20}\underline{72.60} & \cellcolor{grey!20}\underline{88.95} & \cellcolor{grey!20}\underline{72.60} & 88.95 & \cellcolor{grey!20}\underline{72.60} & 88.45 & \cellcolor{grey!20}\underline{72.60} & 89.85 & \cellcolor{grey!20}\underline{72.60} \\
& RepreGuard & \cellcolor{blue!20}\bf{91.70} & \cellcolor{blue!20}\bf{76.30} & \cellcolor{blue!20}\bf{92.00} & \cellcolor{blue!20}\bf{80.40} & \cellcolor{blue!20}\bf{94.25} & \cellcolor{blue!20}\bf{85.30} & \cellcolor{blue!20}\bf{94.95} & \cellcolor{blue!20}\bf{85.10} & \cellcolor{blue!20}\bf{95.05} & \cellcolor{blue!20}\bf{85.50} & \cellcolor{blue!20}\bf{95.05} & \cellcolor{blue!20}\bf{84.80} & \cellcolor{grey!20}\underline{95.30} & \cellcolor{blue!20}\bf{82.60} \\
\midrule
\multirow{6}{*}{Claude-instant} 
& Roberta & 51.90 & 0.00 & 62.30 & 0.00 & 50.85 & 0.00 & 62.10 & 0.20 & 74.45 & 9.30 & \cellcolor{grey!20}\underline{84.85} & 43.90 & 83.70 & 66.00 \\
& LRR & \cellcolor{grey!20}\underline{82.50} & 31.50 & 82.15 & 31.50 & \cellcolor{grey!20}\underline{83.50} & 31.50 & 72.75 & 31.50 & 79.95 & 31.50 & 79.00 & 31.50 & 79.05 & 31.50 \\
& DetectGPT & 53.71 & 0.14 & 54.15 & 0.06 & 54.09 & 0.19 & 55.92 & 0.03 & 56.30 & 0.17 & 57.95 & 0.09 & 57.82 & 0.12 \\
& Fast-Detect. & 79.45 & 58.80 & 79.45 & 58.80 & 80.80 & 58.80 & 80.55 & 58.80 & 78.85 & 58.80 & 80.45 & 58.80 & 79.60 & 58.80 \\
& Binoculars & 82.15 & \cellcolor{blue!20}\bf{72.60} & \cellcolor{grey!20}\underline{82.15} & \cellcolor{blue!20}\bf{72.60} & 82.15 & \cellcolor{blue!20}\bf{72.60} & \cellcolor{grey!20}\underline{82.15} & \cellcolor{grey!20}\underline{72.60} & \cellcolor{grey!20}\underline{82.15} & \cellcolor{grey!20}\underline{72.60} & 81.90 & \cellcolor{grey!20}\underline{72.60} & \cellcolor{grey!20}\underline{85.80} & \cellcolor{grey!20}\underline{72.60} \\
& RepreGuard & \cellcolor{blue!20}\bf{92.40} & \cellcolor{grey!20}\underline{63.20} & \cellcolor{blue!20}\bf{93.25} & \cellcolor{grey!20}\underline{72.50} & \cellcolor{blue!20}\bf{94.55} & \cellcolor{grey!20}\underline{69.90} & \cellcolor{blue!20}\bf{94.60} & \cellcolor{blue!20}\bf{76.30} & \cellcolor{blue!20}\bf{95.40} & \cellcolor{blue!20}\bf{77.00} & \cellcolor{blue!20}\bf{94.80} & \cellcolor{blue!20}\bf{77.10} & \cellcolor{blue!20}\bf{94.75} & \cellcolor{blue!20}\bf{77.30} \\
\midrule
\multirow{6}{*}{Mutil-LLMs} 
& Roberta & 57.25 & 0.00 & 50.45 & 1.60 & 59.80 & 0.70 & 59.60 & 1.30 & 64.95 & 3.60 & \cellcolor{grey!20}\underline{93.50} & 34.80 & \cellcolor{grey!20}\underline{95.65} & 46.90 \\
& LRR & 78.25 & 31.50 & 78.25 & 31.50 & 83.30 & 31.50 & 81.90 & 31.50 & 82.35 & 31.50 & 83.50 & 31.50 & 83.50 & 31.50 \\
& DetectGPT & 55.48 & 0.04 & 54.56 & 0.02 & 54.43 & 0.15 & 54.97 & 0.18 & 55.93 & 0.01 & 56.31 & 0.03 & 57.11 & 0.07 \\
& Fast-Detect. & \cellcolor{grey!20}\underline{84.80} & 58.80 & \cellcolor{grey!20}\underline{84.80} & 58.80 & 85.35 & 58.80 & 85.20 & 58.80 & 85.35 & 58.80 & 85.30 & 58.80 & 85.30 & 58.80 \\
& Binoculars & 82.15 & \cellcolor{grey!20}\underline{72.60} & 82.15 & \cellcolor{grey!20}\underline{72.60} & \cellcolor{grey!20}\underline{89.30} & \cellcolor{grey!20}\underline{72.60} & \cellcolor{grey!20}\underline{89.30} & \cellcolor{grey!20}\underline{72.60} & \cellcolor{grey!20}\underline{89.45} & \cellcolor{grey!20}\underline{72.60} & 90.45 & \cellcolor{grey!20}\underline{72.60} & 90.45 & \cellcolor{grey!20}\underline{72.60} \\
& RepreGuard & \cellcolor{blue!20}\bf{94.20} & \cellcolor{blue!20}\bf{87.10} & \cellcolor{blue!20}\bf{95.85} & \cellcolor{blue!20}\bf{86.90} & \cellcolor{blue!20}\bf{95.70} & \cellcolor{blue!20}\bf{81.90} & \cellcolor{blue!20}\bf{94.20} & \cellcolor{blue!20}\bf{79.90} & \cellcolor{blue!20}\bf{95.80} & \cellcolor{blue!20}\bf{79.80} & \cellcolor{blue!20}\bf{96.30} & \cellcolor{blue!20}\bf{81.20} & \cellcolor{blue!20}\bf{96.00} & \cellcolor{blue!20}\bf{80.20} \\
\bottomrule
\end{tabular}
}
\caption{Performance Comparison of RepreGuard on Various Training Data Shots in Terms of AUROC (\%) on 1000 ``HWT-LGT'' Pairs from 4 Different LLMs. The \colorbox{blue!20}{blue background} or \textbf{bold} indicates the best performance and the \colorbox{grey!20}{grey background} or \underline{underline} indicates the second best.}
\label{tab:trap_data}
\end{table*}

The activation token ratio refers to the proportion of tokens selected from end to the beginning in the samples used to collect neural activation representations for HWT and LGT. Each token’s representation is influenced by all preceding tokens in the text, making this parameter critical for optimizing LGT detection. By focusing on the tokens with the most informative representations, the activation token ratio ensures a better balance between signal and noise. The results in \autoref{fig:activation_representation_ratio} indicate that the average AUROC and TPR@0.01 remains relatively stable as the activation token ratio increases, peaking at a ratio of 0.1, after which it stays consistent until approximately 0.6, followed by a sharp decline. This pattern suggests that not all token positions contribute equally to feature modeling. Specifically, the tokens at the end of the text are more distinctive in differentiating between LGT and HWT texts. As the extends toward the beginning of the sentence, noise gradually increases, leading to a decline in performance. Thus, an optimal activation token ratio is essential for balancing the modelling of important detection features while minimizing noise.

\paragraph{Shots of Training Dataset}

We evaluated the impact of the number of samples used to set the detection threshold, as shown in \autoref{tab:trap_data}. The results indicate that RepreGuard is significantly more effective with limited data compared to other detectors. Specifically, RepreGuard demonstrates optimal performance in the 16-shot setting across most training datasets, achieving an average AUROC of 90.21\% and TPR@0.01 of 77.36\%, surpassing the best statistics-based baseline Binoculars by 6.61\% and 4.76\%, respectively. In addition, while the RoBERTa-based classifier achieved an average AUROC of 91.77\%, which is lower than RepreGuard's average AUROC of 94.79\% in the 1024-shot setting, its TPR@0.01 is significantly lower, averaging only 59.1\%. This indicates that the RoBERTa-based classifier struggles to reliably identify positive cases under stringent false positive constraints, highlighting a limitation in its ability to balance sensitivity and specificity in such scenarios. In contrast, other detectors, such as LRR, DetectGPT and Fast-DetectGPT, demonstrate unstable performance, and consistently fail to exceed a 90\% AUROC even with 1024-shot settings. These findings emphasize the strong and robust detection capabilities of RepreGuard, especially when limited data is available for training.

\begin{figure*}[!ht]
    \centering
    \includegraphics[width=0.99\textwidth, trim=0 0 0 0]{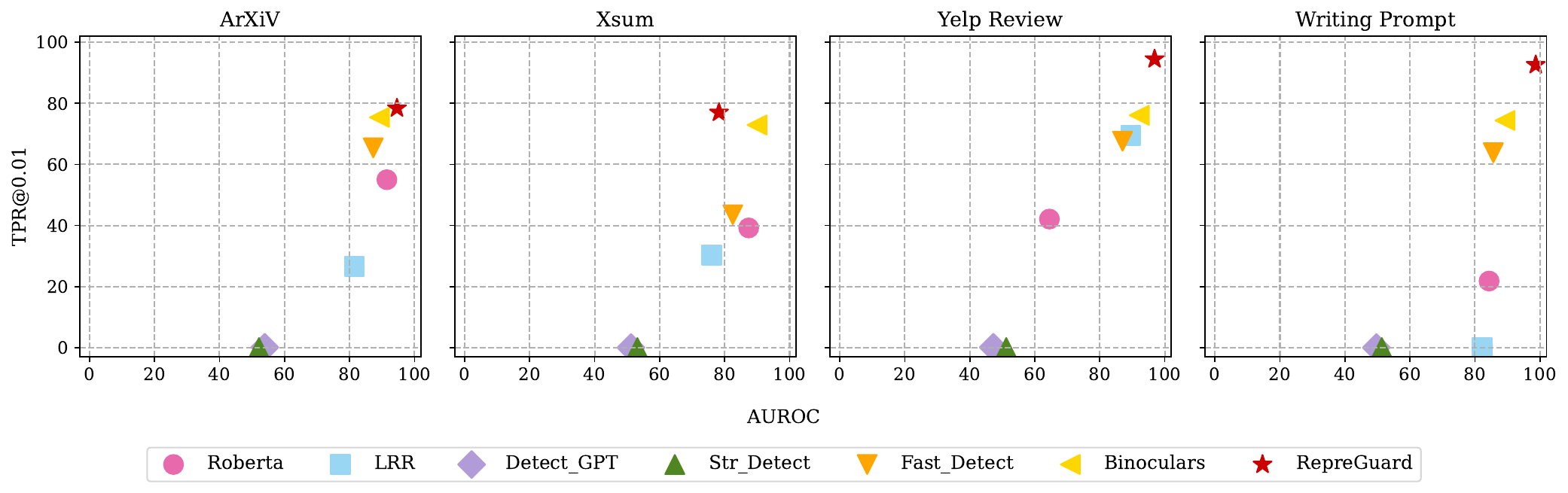}
    \caption{Performance Comparison of Various Detection Methods under OOD Domain Settings across Four Domain in Terms of AUROC (\%) and TPR@0.01 (\%) on a Test Set with 1000 ”HWT-LGT“ Pairs from 4 Different LLMs. The name of each subgraph corresponds to the test domain, while training is conducted on the other three domains. In each domain, the data consists of LGT from four LLMs.}
    \label{fig:domain_performance}
\end{figure*}

\section{Reliability in the Wild}

To evaluate our method in real-world scenarios, we conducted experiments from multiple perspectives, including performance in OOD domains, sensitivity to text length, robustness against paraphrase and perturbation attacks, and the impact of different sampling strategies.
\label{sec:realiabilty}

\subsection{Generalization on Domains}

We evaluated our method on four domain datasets derived from different sources: ArXiv, XSum, Writing Prompt, and Yelp Review. As illustrated in \autoref{fig:domain_performance}, most detectors exhibited poor performance in the OOD domain setting, especially with a low TPR@0.01. However, RepreGuard consistently achieves the highest average AUROC and TPR@0.01 on the OOD domain tasks, with 91.60\% and 85.63\%, respectively. Specifically, RepreGuard achieves the best performance in terms of AUROC and TPR@0.01 under the OOD domain settings for the Arxiv, Writing Prompt, and Yelp Review datasets. While its AUROC on the XSum dataset is slightly lower, RepreGuard still attains the highest TPR@0.01. In contrast, although the Roberta-based classifier, LRR, Fast-DetectGPT and Binoculars demonstrate strong performance in terms of AUROC, their TPR@0.01 values are quite low. For instance, the second-best detector, Binocular, achieved an average AUROC of 88.82\% but a TPR@0.01 of 76.21\%. This indicates that these detectors struggle to identify positive samples when operating at extremely low false positive rates, resulting in a higher risk of misdetections.

\subsection{Detecting Texts with Varied Sizes}
\begin{figure*}[!ht]
    \centering    \includegraphics[width=0.97\textwidth, trim=0 0 0 0]{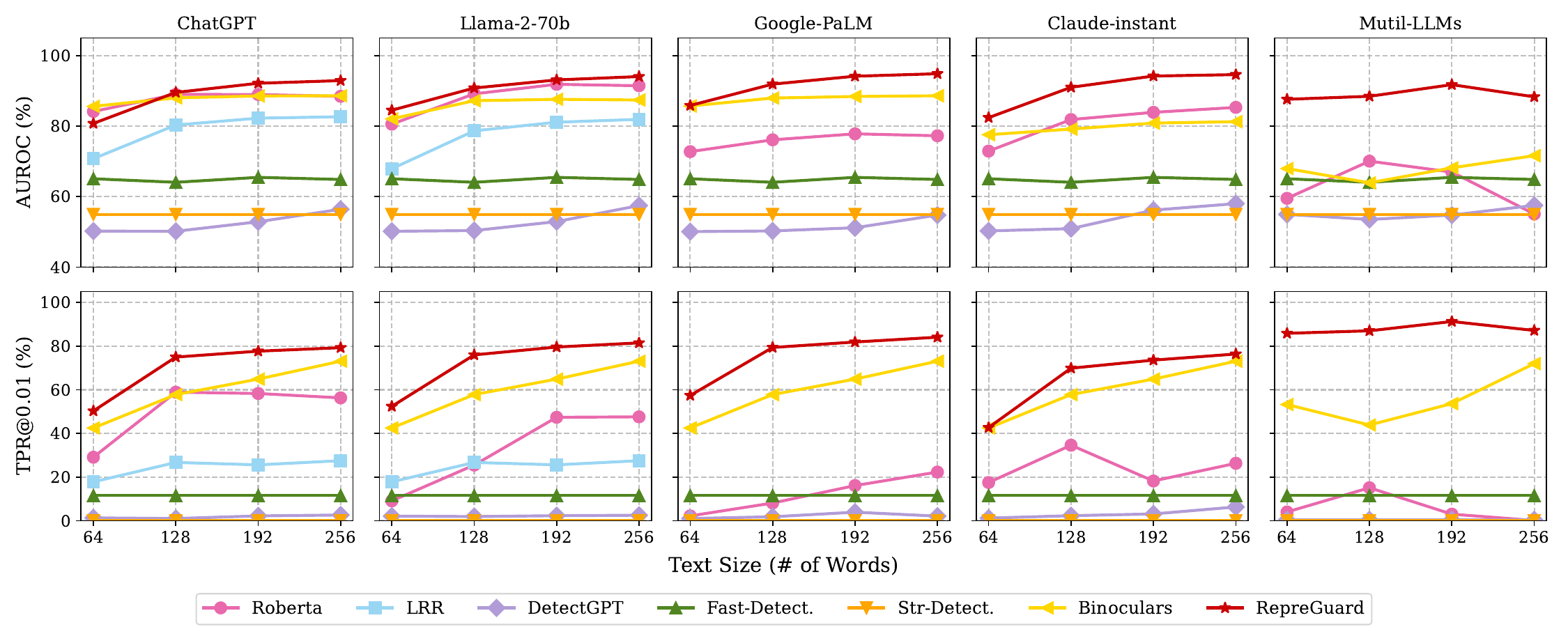}
    \caption{Performance Comparison of RepreGuard on Texts with Varied Sizes in Terms of AUROC (\%) and TPR@0.01 (\%) on a Test Set with 1000 “HWT-LGT” Pairs from 4 Different LLMs. The model name on each sub-graph refers to the LGT from different models used for representation features modeling and threshold setting.}
    \label{fig:Text_Size}
\end{figure*}

We evaluate the impact of text size on the performance of our detector. The results are shown in \autoref{fig:Text_Size}. Overall, RepreGuard achieved the best performance on both short and long texts. Specifically, it attained an AUROC of 84.22\% and a TPR@0.01 of 57.74\% on short texts (64 tokens), while achieving an AUROC of 92.94\% and a TPR@0.01 of 81.70\% on long texts (256 tokens). Although RepreGuard demonstrates slightly lower AUROC performance on 64-token texts compared to other detectors when trained on ChatGPT dataset, its TPR@0.01 consistently outperforms that of other detectors. Furthermore, as the text length increases, the performance advantage of RepreGuard becomes increasingly evident. On 256-token texts, its AUROC and TPR@0.01 significantly surpass those of other detectors, showcasing its exceptional capability in handling long texts. This indicates that RepreGuard remains effective in accurately identifying HWT, minimizing the risk of misclassifying it as LGT, even with shorter text sizes.

\subsection{Robustness on Paraphrase \& Perturbation Attack}

\begin{figure}[!ht]
    \centering    \includegraphics[width=0.5\textwidth, trim=0 0 0 0]{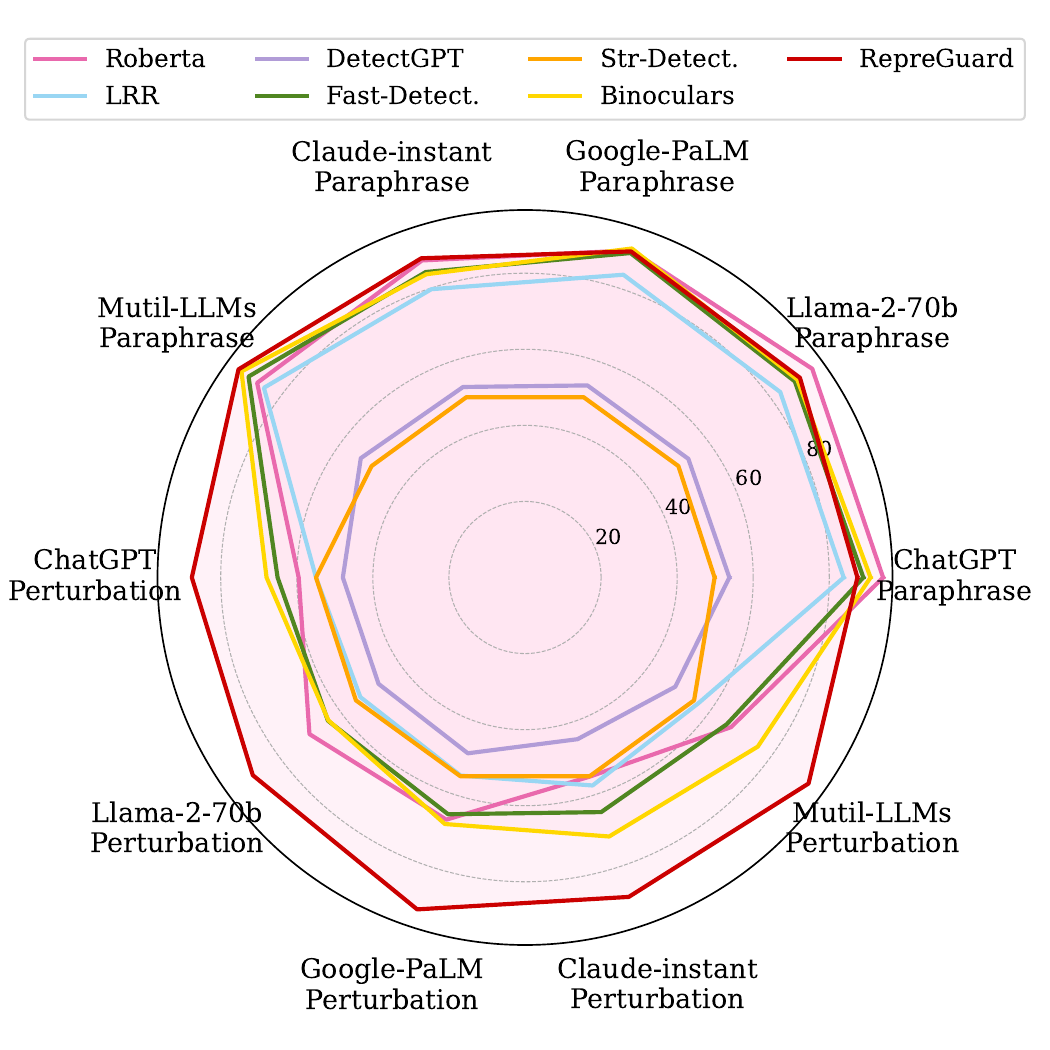}
    \caption{Performance Comparison of RepreGuard on Paraphrase and Perturbation Attack in Terms of AUROC on 1000 “HWT-LGT” Pairs from 4 Different LLMs. The raw text generated by each model is used to model representation features and set thresholds.}
    \label{fig:attacks_auroc}
\end{figure}

We also evaluate the robustness of RepreGuard on mainstream attack methods, including paraphrase attacks and adversarial perturbation attacks. In practical applications, humans often make semantically equivalent revisions to LGT in line with their preferences. In addition, humans might intentionally introduce adversarial noise into LGT to evade detection, creating challenges for the detector.. We used DIPPER Paraphraser~\cite{DBLP:conf/nips/KrishnaSKWI23} and TextBugger~\cite{DBLP:conf/ndss/LiJDLW19} to simulate these realistic scenarios, respectively. The results presented in \autoref{fig:attacks_auroc} and \autoref{fig:attacks_tpr} (see Appendix~\ref{app:attack_tpr}) demonstrate that RepreGuard is the most robust detection method against both paraphrase and perturbation attacks. Significantly, this phenomenon is particularly evident under perturbation attacks, where the AUROC and TPR@0.01 reach 89.65\% and 88.63\%, respectively, exceeding the second-best detector, Binocular, which achieves 69.45\% and 58.54\%. Additionally, although Roberta classifier performs well in certain aspects of AUROC, its TPR@0.01 is extremely poor, dropping as low as 0.10\%, which highlights its significant limitations in identifying positive samples under strict thresholds. In contrast, while certain detectors exhibit strong performance on specific datasets and attacks, such as Fast-DetectGPT achieving an AUROC of 89.70\% on the Google-PaLM dataset under the perturbation attack, and LRR attaining an AUROC of 83.80\% in paragraph attacks on the ChatGPT dataset, their overall performance remains poor, indicating that their robustness is significantly compromised.

\begin{table*}[ht]
    \small
    \centering
    \resizebox{0.98\textwidth}{!}{ 
    \begin{tabular}{l cc cc cc cc cc cc cc cc cc}
    \toprule
    & \multicolumn{8}{c}{\textbf{Chat Models}} & \multicolumn{8}{c}{\textbf{Non-Chat Models}} &  \multicolumn{2}{c}{\textbf{Avg.}} \\
    & \multicolumn{8}{m{10cm}}{\centering (llama-chat, mistral-chat, mpt-chat)} 
    & \multicolumn{8}{m{10cm}}{\centering (mistral, mpt, gpt2)} \\
    \midrule
    \bf Dec. Strategy & \multicolumn{4}{c}{greedy} & \multicolumn{4}{c}{sampling} & \multicolumn{4}{c}{greedy} & \multicolumn{4}{c}{sampling} \\
    \midrule
    \bf Rep. Penalty? & \multicolumn{2}{c}{\xmark} & \multicolumn{2}{c}{\cmark} & \multicolumn{2}{c}{\xmark} & \multicolumn{2}{c}{\cmark} & \multicolumn{2}{c}{\xmark} & \multicolumn{2}{c}{\cmark} & \multicolumn{2}{c}{\xmark} & \multicolumn{2}{c}{\cmark} \\
    \midrule
    \bf Metrics & \textit{AUR.} & \textit{TPR.} & \textit{AUR.} & \textit{TPR.} & \textit{AUR.} & \textit{TPR.} & \textit{AUR.} & \textit{TPR.} & \textit{AUR.} & \textit{TPR.} & \textit{AUR.} & \textit{TPR.} & \textit{AUR.} & \textit{TPR.} & \textit{AUR.} & \textit{TPR.} & \textit{AUR.} & \textit{TPR.} \\
    \midrule
    Roberta & 88.32 & 65.97 & 83.58 & 41.02 & 88.57 & 46.81 & 71.66 & 13.27 & 93.11 & 40.22 & 74.45 & 14.87 & 80.24 & \underline{34.03} & \underline{77.45} & \underline{5.69} & 82.17 & 32.74  \\
    LRR & 90.17 & 49.00 & 80.23 & 12.69 & 86.13 & 24.25 & 67.17 & 4.19 & 94.86 & 87.43 & 83.43 & 34.33 & 77.25 & 0.40 & 50.00 & 0.20 & 78.66 & 26.56 \\
    Fast-Detect. & 97.80 & 95.21 & 87.56 & 68.03 & 97.06 & 91.12 & 71.86 & 30.04 & \underline{98.65} & \underline{96.21} & 77.50 & \underline{48.10} & \underline{84.03} & 31.44 & 50.00 & 0.00 & 83.06 & \underline{57.52} \\
    Str-Detect. & 55.99 & 0.01 & 55.14 & 0.01 & 55.24 & 0.01 & 55.34 & 0.01 & 56.94 & 0.01 & 53.84 & 0.01 & 56.64 & 0.01 & 55.69 & 0.01 & 56.98 & 0.01  \\
    Binoculars & \bf{99.50} & \bf{98.70} & \underline{91.52} & \underline{71.26} & \bf{99.15} & \underline{94.41} & \underline{77.69} & \underline{31.24} & \bf{99.50} & \bf{99.30} & \underline{79.54} & 33.43 & \bf{88.12} & 1.70 & 50.05 & 0.00 & \underline{85.63} & 53.76 \\
    RepreGuard & \underline{98.30} & \underline{96.61} & \bf{97.16} & \bf{94.81} & \underline{97.55} & \bf{94.61} & \bf{94.86} & \bf{85.73} & 98.50 & 92.22 & \bf{92.47} & \bf{75.55} & 72.55 & \bf{34.43} & \bf{81.99} & \bf{46.31} & \bf{92.05} & \bf{77.53} \\
    \bottomrule
    \end{tabular}
        }
    \caption{AUROC and TPR@0.01 for All Detectors Across Model Groups and Sampling Strategies. Sampling with a repetition penalty consistently makes most detectors difficult to detect, while RepreGuard maintains the best performance. The \textbf{Bold} indicates the best performance and \underline{underline} indicates the second best.}
    \label{tab:decoding_full_results}
\end{table*}

\subsection{Various Sampling Methods}
\label{app:appendix_sampling}
\citeauthor{DBLP:conf/iclr/HoltzmanBDFC20} pointed out that sampling strategies with maximum likelihood (such as beam search) often lead to text degeneration. Nucleus sampling addresses these issues by dynamically truncating the long tail of the probability distribution and sampling only from the “nucleus”. This approach effectively avoids degeneration, producing higher-quality and more diverse text, thereby making LGT closer to HWT. To investigate whether different sampling strategies would impact RepreGuard, we utilized the RAID dataset \cite{DBLP:conf/acl/DuganHTZLXIC24} to evaluate the robustness of various sampling strategies. This dataset encompasses multiple domains and generative models and was constructed using diverse sampling approaches. Following the RAID setting, we also divided the data into Chat Models and Non-Chat Models and evaluated the AUROC metric. The results on \autoref{tab:decoding_full_results} demonstrate that RepreGuard achieves the best performance across both models under the different sampling strategies, with an average of 6.42\% in AUROC and 23.77\% in TPR@0.01 higher than Binoculars. Noteworthily, most detectors, like LRR, Fast-Detect. and Binoculars perform well when the repetition penalty mechanism is disabled. However, their AUROC showed a significant decline under the setting of the repetition penalty, whereas RepreGuard demonstrates strong robustness, with its performance only slightly decreasing and even improving on non-chat models. In contrast, most detectors experience a significant performance drop after enabling the penalty mechanism, especially in the sampling scenario of Non-Chat Models, where their detection capability almost completely deteriorates (AUROC approaching 50\%). These indicate that RepreGuard can effectively detect the diversity and complexity of LGT by capturing internal representations, whereas LRR, Fast-Detect and Binoculars only capture information based on output probabilities, leading to the uncertainty introduced by different sampling strategies.

\subsection{Costs of Space and Time}

\begin{table}[ht]
\centering
\resizebox{0.5\textwidth}{!}{
\begin{tabular}{l|ccccc}
\toprule
\bf Detector $\downarrow$ & \textit{AUR.} & \textit{TPR.} & Cost of Space & Cost of Time (Per sample) \\
\midrule
Roberta & 84.85 & 43.90 & \bf{2.0GB} & \bf{0.016s} \\
Fast-Detect. & 80.45 & 58.80 & 40.0GB & 0.390s \\
Binocular & 81.90 & \underline{72.60} & 58.0GB & 0.653s \\
RepreGuard (Phi-2) & \bf{96.10} & 54.50 & \underline{16.0GB} & \underline{0.072s} \\
RepreGuard (Llama-3.1-8B) & \underline{94.80} & \bf{77.10} & 38.0GB & 0.359s \\
\bottomrule
\end{tabular}
}
\caption{Comparison of Effectiveness and Resource Costs on A Test Set with 1,000 “HWT-LGT” Pairs from Four Different LLMs. These was trained on the Claude-Instant dataset with 512 'HWT-LGT' pairs under the setting of NVIDIA A100 80GB using Float32 Precision. The  \textbf{bold} indicates the best performance and \underline{underline} indicates the second best.}
\label{tab:anal_cost_result}
\end{table}

When examining the costs of Methodology, we particularly focus on their balance between effectiveness and resource consumption in real-world applications. To evaluate our method in terms of effectiveness and resource cost, we compared RepreGuard with three other detectors: Roberta, Fast-Detect. and Binocular. In the comparative experiments, we set the batch size to 1 to measure the performance of each method when processing a single sample in the setting of float32 under the A100 80GB GPU. The results in \autoref{tab:anal_cost_result} shown that RepreGuard demonstrates the best overall performance with relatively low resource consumption. Specifically, RepreGuard (Phi-2) achieves the highest AUROC of 96.10\% and relatively low resource consumption (16.0 GB, 0.072 seconds per sample), striking an effective balance between accuracy and efficiency. Meanwhile, RepreGuard (Llama-3.1-8B) achieves an AUROC of 94.80 and the highest TPR@0.01 of 77.10\%, showcasing exceptional capability in positive case detection. It is noteworthy that Roberta lies in its extremely low resource consumption (2.0 GB, 0.016 seconds per sample), making it suitable for cost-constrained scenarios. However, its detection performance (84.85\% in AUROC, 43.90\% in TPR@0.01) is significantly inferior to that of RepreGuard.

In addition, we assess whether our approach is affected
by memorization (see Appendix~\ref{app:analysis_of_model_memorization}), and
examine how the performance as the increase of
the LGT used in LLMs (see Appendix~\ref{app:performance_after_aigc_pretraining})

\section{Conclusion}

In this paper, we introduce \textbf{RepreGuard}, a novel and reliable method based on hidden representation features for detecting text generated by LLMs. Experimental results on both ID and OOD, demonstrate RepreGuard's strong detection capabilities and zero-shot proficiency. It requires only a small number of training samples to achieve impressive OOD generalization, effectively handling diverse real-world application scenarios and challenge from newly emerging LLMs. Furthermore, we verify the effectiveness, robustness, and generalization ability of RepreGuard in detecting texts of varied sizes, as well as texts that have undergone paraphrasing attack, perturbation attack, and various sampling methods.

\section*{Acknowledgments}

This work was supported in part by the Science and Technology Development Fund of Macau SAR (Grant No. FDCT/0007/2024/AKP), the Science and Technology Development Fund of Macau SAR (Grant No. FDCT/0070/2022/AMJ, China Strategic Scientific and Technological Innovation Cooperation Project Grant No. 2022YFE0204900), the Science and Technology Development Fund of Macau SAR (Grant No. FDCT/060/2022/AFJ, National Natural Science Foundation of China Grant No. 62261160648), the UM and UMDF (Grant Nos. MYRG-GRG2023-00006-FST-UMDF, MYRG-GRG2024-00165-FST-UMDF, EF2024-00185-FST), and the National Natural Science Foundation of China (Grant No. 62266013). This work was also supported in part by National Key Research and Development Program of China (2022YFF0902100), National Natural Science Foundation of China (Grant No. 62376262), the Natural Science Foundation of Guangdong Province of China (2024A1515030166, 2025B1515020032), and the Shenzhen Science and Technology Innovation Program (KQTD20190929172835662).

\bibliography{tacl2021}

\begin{thebibliography}{51}
\expandafter\ifx\csname natexlab\endcsname\relax\def\natexlab#1{#1}\fi

\bibitem[{Anthropic(2023)}]{blog2024claude_instant}
Anthropic. 2023.
\newblock \href {https://www.anthropic.com/news/releasing-claude-instant-1-2} {Releasing claude instant 1.2}.

\bibitem[{Azaria and Mitchell(2023)}]{azaria-mitchell-2023-internal}
Amos Azaria and Tom Mitchell. 2023.
\newblock \href {https://doi.org/10.18653/v1/2023.findings-emnlp.68} {The internal state of an {LLM} knows when it`s lying}.
\newblock In \emph{Findings of the Association for Computational Linguistics: EMNLP 2023}, pages 967--976, Singapore. Association for Computational Linguistics.

\bibitem[{Bao et~al.(2024)Bao, Zhao, Teng, Yang, and Zhang}]{DBLP:conf/iclr/BaoZTY024}
Guangsheng Bao, Yanbin Zhao, Zhiyang Teng, Linyi Yang, and Yue Zhang. 2024.
\newblock \href {https://openreview.net/forum?id=Bpcgcr8E8Z} {Fast-detectgpt: Efficient zero-shot detection of machine-generated text via conditional probability curvature}.
\newblock In \emph{The Twelfth International Conference on Learning Representations, {ICLR} 2024, Vienna, Austria, May 7-11, 2024}.

\bibitem[{Black et~al.(2021)Black, Gao, Wang, Leahy, and Biderman}]{gpt-neo}
Sid Black, Leo Gao, Phil Wang, Connor Leahy, and Stella Biderman. 2021.
\newblock \href {https://doi.org/10.5281/zenodo.5297715} {Gpt-neo: Large scale autoregressive language modeling with mesh-tensorflow}.

\bibitem[{Carlini et~al.(2021)Carlini, Tram{\`{e}}r, Wallace, Jagielski, Herbert{-}Voss, Lee, Roberts, Brown, Song, Erlingsson, Oprea, and Raffel}]{DBLP:conf/uss/CarliniTWJHLRBS21}
Nicholas Carlini, Florian Tram{\`{e}}r, Eric Wallace, Matthew Jagielski, Ariel Herbert{-}Voss, Katherine Lee, Adam Roberts, Tom~B. Brown, Dawn Song, {\'{U}}lfar Erlingsson, Alina Oprea, and Colin Raffel. 2021.
\newblock \href {https://www.usenix.org/conference/usenixsecurity21/presentation/carlini-extracting} {Extracting training data from large language models}.
\newblock In \emph{30th {USENIX} Security Symposium, {USENIX} Security 2021, August 11-13, 2021}, pages 2633--2650. {USENIX} Association.

\bibitem[{Chen et~al.(2023)Chen, Kang, Zhai, Li, Singh, and Raj}]{DBLP:journals/corr/abs-2305-07969}
Yutian Chen, Hao Kang, Vivian Zhai, Liangze Li, Rita Singh, and Bhiksha Raj. 2023.
\newblock \href {https://doi.org/10.48550/ARXIV.2305.07969} {Gpt-sentinel: Distinguishing human and chatgpt generated content}.
\newblock \emph{CoRR}, abs/2305.07969.

\bibitem[{Cotton et~al.(2024)Cotton, Cotton, and Shipway}]{cotton2024chatting}
Debby~RE Cotton, Peter~A Cotton, and J~Reuben Shipway. 2024.
\newblock Chatting and cheating: Ensuring academic integrity in the era of chatgpt.
\newblock \emph{Innovations in education and teaching international}, 61(2):228--239.

\bibitem[{Dugan et~al.(2024)Dugan, Hwang, Trhl{\'{\i}}k, Zhu, Ludan, Xu, Ippolito, and Callison{-}Burch}]{DBLP:conf/acl/DuganHTZLXIC24}
Liam Dugan, Alyssa Hwang, Filip Trhl{\'{\i}}k, Andrew Zhu, Josh~Magnus Ludan, Hainiu Xu, Daphne Ippolito, and Chris Callison{-}Burch. 2024.
\newblock \href {https://doi.org/10.18653/V1/2024.ACL-LONG.674} {{RAID:} {A} shared benchmark for robust evaluation of machine-generated text detectors}.
\newblock In \emph{Proceedings of the 62nd Annual Meeting of the Association for Computational Linguistics (Volume 1: Long Papers), {ACL} 2024, Bangkok, Thailand, August 11-16, 2024}, pages 12463--12492. Association for Computational Linguistics.

\bibitem[{Durrani et~al.(2020)Durrani, Sajjad, Dalvi, and Belinkov}]{DBLP:conf/emnlp/DurraniSDB20}
Nadir Durrani, Hassan Sajjad, Fahim Dalvi, and Yonatan Belinkov. 2020.
\newblock \href {https://doi.org/10.18653/V1/2020.EMNLP-MAIN.395} {Analyzing individual neurons in pre-trained language models}.
\newblock In \emph{Proceedings of the 2020 Conference on Empirical Methods in Natural Language Processing, {EMNLP} 2020, Online, November 16-20, 2020}, pages 4865--4880. Association for Computational Linguistics.

\bibitem[{Fagni et~al.(2020)Fagni, Falchi, Gambini, Martella, and Tesconi}]{DBLP:journals/corr/abs-2008-00036}
Tiziano Fagni, Fabrizio Falchi, Margherita Gambini, Antonio Martella, and Maurizio Tesconi. 2020.
\newblock \href {http://arxiv.org/abs/2008.00036} {Tweepfake: about detecting deepfake tweets}.
\newblock \emph{CoRR}, abs/2008.00036.

\bibitem[{Fan et~al.(2018)Fan, Lewis, and Dauphin}]{DBLP:conf/acl/LewisDF18}
Angela Fan, Mike Lewis, and Yann~N. Dauphin. 2018.
\newblock \href {https://doi.org/10.18653/V1/P18-1082} {Hierarchical neural story generation}.
\newblock In \emph{Proceedings of the 56th Annual Meeting of the Association for Computational Linguistics, {ACL} 2018, Melbourne, Australia, July 15-20, 2018, Volume 1: Long Papers}, pages 889--898.

\bibitem[{Ghahramani(2023)}]{blog2023palm}
Zoubin Ghahramani. 2023.
\newblock \href {https://blog.google/technology/ai/google-palm-2-ai-large-language-model/} {Introducing palm 2}.

\bibitem[{Guo et~al.(2023)Guo, Zhang, Wang, Jiang, Nie, Ding, Yue, and Wu}]{DBLP:journals/corr/abs-2301-07597}
Biyang Guo, Xin Zhang, Ziyuan Wang, Minqi Jiang, Jinran Nie, Yuxuan Ding, Jianwei Yue, and Yupeng Wu. 2023.
\newblock \href {https://doi.org/10.48550/ARXIV.2301.07597} {How close is chatgpt to human experts? comparison corpus, evaluation, and detection}.
\newblock \emph{CoRR}, abs/2301.07597.

\bibitem[{Hans et~al.(2024)Hans, Schwarzschild, Cherepanova, Kazemi, Saha, Goldblum, Geiping, and Goldstein}]{DBLP:conf/icml/HansSCKSGGG24}
Abhimanyu Hans, Avi Schwarzschild, Valeriia Cherepanova, Hamid Kazemi, Aniruddha Saha, Micah Goldblum, Jonas Geiping, and Tom Goldstein. 2024.
\newblock \href {https://openreview.net/forum?id=axl3FAkpik} {Spotting llms with binoculars: Zero-shot detection of machine-generated text}.
\newblock In \emph{Forty-first International Conference on Machine Learning, {ICML} 2024, Vienna, Austria, July 21-27, 2024}.

\bibitem[{Holtzman et~al.(2020)Holtzman, Buys, Du, Forbes, and Choi}]{DBLP:conf/iclr/HoltzmanBDFC20}
Ari Holtzman, Jan Buys, Li~Du, Maxwell Forbes, and Yejin Choi. 2020.
\newblock \href {https://openreview.net/forum?id=rygGQyrFvH} {The curious case of neural text degeneration}.
\newblock In \emph{8th International Conference on Learning Representations, {ICLR} 2020, Addis Ababa, Ethiopia, April 26-30, 2020}. OpenReview.net.

\bibitem[{Ji et~al.(2023)Ji, Lee, Frieske, Yu, Su, Xu, Ishii, Bang, Madotto, and Fung}]{DBLP:journals/csur/JiLFYSXIBMF23}
Ziwei Ji, Nayeon Lee, Rita Frieske, Tiezheng Yu, Dan Su, Yan Xu, Etsuko Ishii, Yejin Bang, Andrea Madotto, and Pascale Fung. 2023.
\newblock \href {https://doi.org/10.1145/3571730} {Survey of hallucination in natural language generation}.
\newblock \emph{{ACM} Comput. Surv.}, 55(12):248:1--248:38.

\bibitem[{Krishna et~al.(2023)Krishna, Song, Karpinska, Wieting, and Iyyer}]{DBLP:conf/nips/KrishnaSKWI23}
Kalpesh Krishna, Yixiao Song, Marzena Karpinska, John Wieting, and Mohit Iyyer. 2023.
\newblock \href {http://papers.nips.cc/paper\_files/paper/2023/hash/575c450013d0e99e4b0ecf82bd1afaa4-Abstract-Conference.html} {Paraphrasing evades detectors of ai-generated text, but retrieval is an effective defense}.
\newblock In \emph{Advances in Neural Information Processing Systems 36: Annual Conference on Neural Information Processing Systems 2023, NeurIPS 2023, New Orleans, LA, USA, December 10 - 16, 2023}.

\bibitem[{Li et~al.(2019)Li, Ji, Du, Li, and Wang}]{DBLP:conf/ndss/LiJDLW19}
Jinfeng Li, Shouling Ji, Tianyu Du, Bo~Li, and Ting Wang. 2019.
\newblock \href {https://www.ndss-symposium.org/ndss-paper/textbugger-generating-adversarial-text-against-real-world-applications/} {Textbugger: Generating adversarial text against real-world applications}.
\newblock In \emph{26th Annual Network and Distributed System Security Symposium, {NDSS} 2019, San Diego, California, USA, February 24-27, 2019}.

\bibitem[{Liu et~al.(2023)Liu, Zhang, Zhang, Yue, Zhao, Cheng, Zhang, and Hu}]{DBLP:journals/corr/abs-2304-07666}
Yikang Liu, Ziyin Zhang, Wanyang Zhang, Shisen Yue, Xiaojing Zhao, Xinyuan Cheng, Yiwen Zhang, and Hai Hu. 2023.
\newblock \href {https://doi.org/10.48550/ARXIV.2304.07666} {Argugpt: evaluating, understanding and identifying argumentative essays generated by {GPT} models}.
\newblock \emph{CoRR}, abs/2304.07666.

\bibitem[{Liu et~al.(2019)Liu, Ott, Goyal, Du, Joshi, Chen, Levy, Lewis, Zettlemoyer, and Stoyanov}]{DBLP:journals/corr/abs-1907-11692}
Yinhan Liu, Myle Ott, Naman Goyal, Jingfei Du, Mandar Joshi, Danqi Chen, Omer Levy, Mike Lewis, Luke Zettlemoyer, and Veselin Stoyanov. 2019.
\newblock \href {http://arxiv.org/abs/1907.11692} {Roberta: {A} robustly optimized {BERT} pretraining approach}.
\newblock \emph{CoRR}, abs/1907.11692.

\bibitem[{MetaAI(2024)}]{blog2024llama3}
MetaAI. 2024.
\newblock \href {https://ai.meta.com/blog/meta-llama-3/} {Introducing meta llama 3: The most capable openly available llm to date}.

\bibitem[{Mitchell et~al.(2023)Mitchell, Lee, Khazatsky, Manning, and Finn}]{DBLP:conf/icml/Mitchell0KMF23}
Eric Mitchell, Yoonho Lee, Alexander Khazatsky, Christopher~D. Manning, and Chelsea Finn. 2023.
\newblock \href {https://proceedings.mlr.press/v202/mitchell23a.html} {Detectgpt: Zero-shot machine-generated text detection using probability curvature}.
\newblock In \emph{International Conference on Machine Learning, {ICML} 2023, 23-29 July 2023, Honolulu, Hawaii, {USA}}, pages 24950--24962.

\bibitem[{Monea et~al.(2024)Monea, Peyrard, Josifoski, Chaudhary, Eisner, Kiciman, Palangi, Patra, and West}]{monea-etal-2024-glitch}
Giovanni Monea, Maxime Peyrard, Martin Josifoski, Vishrav Chaudhary, Jason Eisner, Emre Kiciman, Hamid Palangi, Barun Patra, and Robert West. 2024.
\newblock \href {https://doi.org/10.18653/v1/2024.acl-long.369} {A glitch in the matrix? locating and detecting language model grounding with fakepedia}.
\newblock In \emph{Proceedings of the 62nd Annual Meeting of the Association for Computational Linguistics (Volume 1: Long Papers)}, pages 6828--6844, Bangkok, Thailand. Association for Computational Linguistics.

\bibitem[{Mukherjee et~al.(2023)Mukherjee, Mitra, Jawahar, Agarwal, Palangi, and Awadallah}]{DBLP:journals/corr/abs-2306-02707}
Subhabrata Mukherjee, Arindam Mitra, Ganesh Jawahar, Sahaj Agarwal, Hamid Palangi, and Ahmed Awadallah. 2023.
\newblock \href {https://doi.org/10.48550/ARXIV.2306.02707} {Orca: Progressive learning from complex explanation traces of {GPT-4}}.
\newblock \emph{CoRR}, abs/2306.02707.

\bibitem[{Narayan et~al.(2018)Narayan, Cohen, and Lapata}]{DBLP:conf/emnlp/NarayanCL18}
Shashi Narayan, Shay~B. Cohen, and Mirella Lapata. 2018.
\newblock \href {https://doi.org/10.18653/V1/D18-1206} {Don't give me the details, just the summary! topic-aware convolutional neural networks for extreme summarization}.
\newblock In \emph{Proceedings of the 2018 Conference on Empirical Methods in Natural Language Processing, Brussels, Belgium, October 31 - November 4, 2018}, pages 1797--1807.

\bibitem[{OpenAI(2022)}]{blog2022chatgpt}
OpenAI. 2022.
\newblock \href {https://openai.com/index/chatgpt/} {Introducing chatgpt}.

\bibitem[{Pagnoni et~al.(2022)Pagnoni, Graciarena, and Tsvetkov}]{DBLP:conf/coling/PagnoniGT22}
Artidoro Pagnoni, Martin Graciarena, and Yulia Tsvetkov. 2022.
\newblock \href {https://aclanthology.org/2022.coling-1.106} {Threat scenarios and best practices to detect neural fake news}.
\newblock In \emph{Proceedings of the 29th International Conference on Computational Linguistics, {COLING} 2022, Gyeongju, Republic of Korea, October 12-17, 2022}, pages 1233--1249.

\bibitem[{Penedo et~al.(2023)Penedo, Malartic, Hesslow, Cojocaru, Alobeidli, Cappelli, Pannier, Almazrouei, and Launay}]{DBLP:conf/nips/PenedoMHCACPAL23}
Guilherme Penedo, Quentin Malartic, Daniel Hesslow, Ruxandra Cojocaru, Hamza Alobeidli, Alessandro Cappelli, Baptiste Pannier, Ebtesam Almazrouei, and Julien Launay. 2023.
\newblock \href {http://papers.nips.cc/paper\_files/paper/2023/hash/fa3ed726cc5073b9c31e3e49a807789c-Abstract-Datasets\_and\_Benchmarks.html} {The refinedweb dataset for falcon {LLM:} outperforming curated corpora with web data only}.
\newblock In \emph{Advances in Neural Information Processing Systems 36: Annual Conference on Neural Information Processing Systems 2023, NeurIPS 2023, New Orleans, LA, USA, December 10 - 16, 2023}.

\bibitem[{Raffel et~al.(2020)Raffel, Shazeer, Roberts, Lee, Narang, Matena, Zhou, Li, and Liu}]{DBLP:journals/jmlr/RaffelSRLNMZLL20}
Colin Raffel, Noam Shazeer, Adam Roberts, Katherine Lee, Sharan Narang, Michael Matena, Yanqi Zhou, Wei Li, and Peter~J. Liu. 2020.
\newblock \href {https://jmlr.org/papers/v21/20-074.html} {Exploring the limits of transfer learning with a unified text-to-text transformer}.
\newblock \emph{J. Mach. Learn. Res.}, 21:140:1--140:67.

\bibitem[{Sarvazyan et~al.(2023)Sarvazyan, Gonz{\'{a}}lez, Rosso, and Franco{-}Salvador}]{DBLP:conf/clef/SarvazyanGRF23}
Areg~Mikael Sarvazyan, Jos{\'{e}}~{\'{A}}ngel Gonz{\'{a}}lez, Paolo Rosso, and Marc Franco{-}Salvador. 2023.
\newblock \href {https://doi.org/10.1007/978-3-031-42448-9\_11} {Supervised machine-generated text detectors: Family and scale matters}.
\newblock In \emph{Experimental {IR} Meets Multilinguality, Multimodality, and Interaction - 14th International Conference of the {CLEF} Association, {CLEF} 2023, Thessaloniki, Greece, September 18-21, 2023, Proceedings}, pages 121--132.

\bibitem[{Solaiman et~al.(2019)Solaiman, Brundage, Clark, Askell, Herbert{-}Voss, Wu, Radford, and Wang}]{DBLP:journals/corr/abs-1908-09203}
Irene Solaiman, Miles Brundage, Jack Clark, Amanda Askell, Ariel Herbert{-}Voss, Jeff Wu, Alec Radford, and Jasmine Wang. 2019.
\newblock \href {http://arxiv.org/abs/1908.09203} {Release strategies and the social impacts of language models}.
\newblock \emph{CoRR}, abs/1908.09203.

\bibitem[{Sriramanan et~al.(2024)Sriramanan, Bharti, Sadasivan, Saha, Kattakinda, and Feizi}]{NEURIPS2024_3c1e1fdf}
Gaurang Sriramanan, Siddhant Bharti, Vinu~Sankar Sadasivan, Shoumik Saha, Priyatham Kattakinda, and Soheil Feizi. 2024.
\newblock \href {https://proceedings.neurips.cc/paper_files/paper/2024/file/3c1e1fdf305195cd620c118aaa9717ad-Paper-Conference.pdf} {Llm-check: Investigating detection of hallucinations in large language models}.
\newblock In \emph{Advances in Neural Information Processing Systems}, volume~37, pages 34188--34216. Curran Associates, Inc.

\bibitem[{Su et~al.(2023)Su, Zhuo, Wang, and Nakov}]{DBLP:conf/emnlp/SuZ0N23}
Jinyan Su, Terry~Yue Zhuo, Di~Wang, and Preslav Nakov. 2023.
\newblock \href {https://doi.org/10.18653/V1/2023.FINDINGS-EMNLP.827} {Detectllm: Leveraging log rank information for zero-shot detection of machine-generated text}.
\newblock In \emph{Findings of the Association for Computational Linguistics: {EMNLP} 2023, Singapore, December 6-10, 2023}, pages 12395--12412.

\bibitem[{Sun et~al.(2025)Sun, Zhang, Shen, Zhang, Liu, Backes, Zhang, and He}]{sun2025aigeneratedtextworldalready}
Zhen Sun, Zongmin Zhang, Xinyue Shen, Ziyi Zhang, Yule Liu, Michael Backes, Yang Zhang, and Xinlei He. 2025.
\newblock \href {http://arxiv.org/abs/2412.18148} {Are we in the ai-generated text world already? quantifying and monitoring aigt on social media}.

\bibitem[{Tang et~al.(2024)Tang, Luo, Huang, Zhang, Wang, Zhao, Wei, and Wen}]{DBLP:conf/acl/TangLH0WZWW24}
Tianyi Tang, Wenyang Luo, Haoyang Huang, Dongdong Zhang, Xiaolei Wang, Xin Zhao, Furu Wei, and Ji{-}Rong Wen. 2024.
\newblock \href {https://doi.org/10.18653/V1/2024.ACL-LONG.309} {Language-specific neurons: The key to multilingual capabilities in large language models}.
\newblock In \emph{Proceedings of the 62nd Annual Meeting of the Association for Computational Linguistics (Volume 1: Long Papers), {ACL} 2024, Bangkok, Thailand, August 11-16, 2024}, pages 5701--5715. Association for Computational Linguistics.

\bibitem[{Tjuatja et~al.(2023)Tjuatja, Chen, Wu, Talwalkar, and Neubig}]{DBLP:journals/corr/abs-2311-04076}
Lindia Tjuatja, Valerie Chen, Sherry~Tongshuang Wu, Ameet Talwalkar, and Graham Neubig. 2023.
\newblock \href {https://doi.org/10.48550/ARXIV.2311.04076} {Do llms exhibit human-like response biases? {A} case study in survey design}.
\newblock \emph{CoRR}, abs/2311.04076.

\bibitem[{Verma et~al.(2024)Verma, Fleisig, Tomlin, and Klein}]{verma-etal-2024-ghostbuster}
Vivek Verma, Eve Fleisig, Nicholas Tomlin, and Dan Klein. 2024.
\newblock \href {https://doi.org/10.18653/v1/2024.naacl-long.95} {Ghostbuster: Detecting text ghostwritten by large language models}.
\newblock In \emph{Proceedings of the 2024 Conference of the North American Chapter of the Association for Computational Linguistics: Human Language Technologies (Volume 1: Long Papers)}, pages 1702--1717, Mexico City, Mexico. Association for Computational Linguistics.

\bibitem[{Voita et~al.(2024)Voita, Ferrando, and Nalmpantis}]{DBLP:conf/acl/VoitaFN24}
Elena Voita, Javier Ferrando, and Christoforos Nalmpantis. 2024.
\newblock \href {https://doi.org/10.18653/V1/2024.FINDINGS-ACL.75} {Neurons in large language models: Dead, n-gram, positional}.
\newblock In \emph{Findings of the Association for Computational Linguistics, {ACL} 2024, Bangkok, Thailand and virtual meeting, August 11-16, 2024}, pages 1288--1301. Association for Computational Linguistics.

\bibitem[{Wang and Komatsuzaki(2021)}]{gpt-j}
Ben Wang and Aran Komatsuzaki. 2021.
\newblock {GPT-J-6B: A 6 Billion Parameter Autoregressive Language Model}.
\newblock \url{https://github.com/kingoflolz/mesh-transformer-jax}.

\bibitem[{Wang et~al.(2023)Wang, Cheng, Cui, and Yu}]{DBLP:journals/corr/abs-2306-07401}
Zecong Wang, Jiaxi Cheng, Chen Cui, and Chenhao Yu. 2023.
\newblock \href {https://doi.org/10.48550/ARXIV.2306.07401} {Implementing {BERT} and fine-tuned roberta to detect {AI} generated news by chatgpt}.
\newblock \emph{CoRR}, abs/2306.07401.

\bibitem[{Wu et~al.(2023)Wu, Yang, Zhan, Yuan, Wong, and Chao}]{DBLP:journals/corr/abs-2310-14724}
Junchao Wu, Shu Yang, Runzhe Zhan, Yulin Yuan, Derek~F. Wong, and Lidia~S. Chao. 2023.
\newblock \href {https://doi.org/10.48550/ARXIV.2310.14724} {A survey on llm-generated text detection: Necessity, methods, and future directions}.
\newblock \emph{CoRR}, abs/2310.14724.

\bibitem[{Wu et~al.(2025)Wu, Zhan, Wong, Yang, Liu, Chao, and Zhang}]{wu2025wrote}
Junchao Wu, Runzhe Zhan, Derek~F Wong, Shu Yang, Xuebo Liu, Lidia~S Chao, and Min Zhang. 2025.
\newblock Who wrote this? the key to zero-shot llm-generated text detection is gecscore.
\newblock In \emph{Proceedings of the 31st International Conference on Computational Linguistics}, pages 10275--10292.

\bibitem[{Wu et~al.(2024)Wu, Zhan, Wong, Yang, Yang, Yuan, and Chao}]{DBLP:conf/nips/WuZWY0YC24}
Junchao Wu, Runzhe Zhan, Derek~F. Wong, Shu Yang, Xinyi Yang, Yulin Yuan, and Lidia~S. Chao. 2024.
\newblock \href {http://papers.nips.cc/paper\_files/paper/2024/hash/b61bdf7e9f64c04ec75a26e781e2ad51-Abstract-Datasets\_and\_Benchmarks\_Track.html} {Detectrl: Benchmarking llm-generated text detection in real-world scenarios}.
\newblock In \emph{Advances in Neural Information Processing Systems 38: Annual Conference on Neural Information Processing Systems 2024, NeurIPS 2024, Vancouver, BC, Canada, December 10 - 15, 2024}.

\bibitem[{Xu et~al.(2024{\natexlab{a}})Xu, Sun, Zheng, Geng, Zhao, Feng, Tao, Lin, and Jiang}]{DBLP:conf/iclr/XuSZG0FTLJ24}
Can Xu, Qingfeng Sun, Kai Zheng, Xiubo Geng, Pu~Zhao, Jiazhan Feng, Chongyang Tao, Qingwei Lin, and Daxin Jiang. 2024{\natexlab{a}}.
\newblock \href {https://openreview.net/forum?id=CfXh93NDgH} {Wizardlm: Empowering large pre-trained language models to follow complex instructions}.
\newblock In \emph{The Twelfth International Conference on Learning Representations, {ICLR} 2024, Vienna, Austria, May 7-11, 2024}. OpenReview.net.

\bibitem[{Xu et~al.(2024{\natexlab{b}})Xu, Zhan, Wong, and Chao}]{DBLP:journals/corr/abs-2403-11621}
Haoyun Xu, Runzhe Zhan, Derek~F. Wong, and Lidia~S. Chao. 2024{\natexlab{b}}.
\newblock \href {https://doi.org/10.48550/ARXIV.2403.11621} {Let's focus on neuron: Neuron-level supervised fine-tuning for large language model}.
\newblock \emph{CoRR}, abs/2403.11621.

\bibitem[{Yang et~al.(2024)Yang, Cheng, Wu, Petzold, Wang, and Chen}]{DBLP:conf/iclr/Yang0WPWC24}
Xianjun Yang, Wei Cheng, Yue Wu, Linda~Ruth Petzold, William~Yang Wang, and Haifeng Chen. 2024.
\newblock \href {https://openreview.net/forum?id=Xlayxj2fWp} {{DNA-GPT:} divergent n-gram analysis for training-free detection of gpt-generated text}.
\newblock In \emph{The Twelfth International Conference on Learning Representations, {ICLR} 2024, Vienna, Austria, May 7-11, 2024}.

\bibitem[{Yu et~al.(2023)Yu, Pang, Liu, Du, Kang, Huang, Lin, and Yan}]{DBLP:conf/icml/YuPLDKHLY23}
Weichen Yu, Tianyu Pang, Qian Liu, Chao Du, Bingyi Kang, Yan Huang, Min Lin, and Shuicheng Yan. 2023.
\newblock \href {https://proceedings.mlr.press/v202/yu23c.html} {Bag of tricks for training data extraction from language models}.
\newblock In \emph{International Conference on Machine Learning, {ICML} 2023, 23-29 July 2023, Honolulu, Hawaii, {USA}}, volume 202 of \emph{Proceedings of Machine Learning Research}, pages 40306--40320. {PMLR}.

\bibitem[{Zellers et~al.(2019)Zellers, Holtzman, Rashkin, Bisk, Farhadi, Roesner, and Choi}]{DBLP:conf/nips/ZellersHRBFRC19}
Rowan Zellers, Ari Holtzman, Hannah Rashkin, Yonatan Bisk, Ali Farhadi, Franziska Roesner, and Yejin Choi. 2019.
\newblock \href {https://proceedings.neurips.cc/paper/2019/hash/3e9f0fc9b2f89e043bc6233994dfcf76-Abstract.html} {Defending against neural fake news}.
\newblock In \emph{Advances in Neural Information Processing Systems 32: Annual Conference on Neural Information Processing Systems 2019, NeurIPS 2019, December 8-14, 2019, Vancouver, BC, Canada}, pages 9051--9062.

\bibitem[{Zhang et~al.(2015)Zhang, Zhao, and LeCun}]{DBLP:conf/nips/ZhangZL15}
Xiang Zhang, Junbo~Jake Zhao, and Yann LeCun. 2015.
\newblock \href {https://proceedings.neurips.cc/paper/2015/hash/250cf8b51c773f3f8dc8b4be867a9a02-Abstract.html} {Character-level convolutional networks for text classification}.
\newblock In \emph{Advances in Neural Information Processing Systems 28: Annual Conference on Neural Information Processing Systems 2015, December 7-12, 2015, Montreal, Quebec, Canada}, pages 649--657.

\bibitem[{Zhang et~al.(2024)Zhang, Hu, Lee, Shi, Kordjamshidi, Chai, and Ma}]{zhang2025do}
Zheyuan Zhang, Fengyuan Hu, Jayjun Lee, Freda Shi, Parisa Kordjamshidi, Joyce Chai, and Ziqiao Ma. 2024.
\newblock \href {https://openreview.net/forum?id=84pDoCD4lH} {Do vision-language models represent space and how? evaluating spatial frame of reference under ambiguities}.
\newblock In \emph{The Thirteenth International Conference on Learning Representations}.

\bibitem[{Zou et~al.(2023)Zou, Phan, Chen, Campbell, Guo, Ren, Pan, Yin, Mazeika, Dombrowski, Goel, Li, Byun, Wang, Mallen, Basart, Koyejo, Song, Fredrikson, Kolter, and Hendrycks}]{DBLP:journals/corr/abs-2310-01405}
Andy Zou, Long Phan, Sarah Chen, James Campbell, Phillip Guo, Richard Ren, Alexander Pan, Xuwang Yin, Mantas Mazeika, Ann{-}Kathrin Dombrowski, Shashwat Goel, Nathaniel Li, Michael~J. Byun, Zifan Wang, Alex Mallen, Steven Basart, Sanmi Koyejo, Dawn Song, Matt Fredrikson, J.~Zico Kolter, and Dan Hendrycks. 2023.
\newblock \href {https://doi.org/10.48550/ARXIV.2310.01405} {Representation engineering: {A} top-down approach to {AI} transparency}.
\newblock \emph{CoRR}, abs/2310.01405.

\end{thebibliography}
\bibliographystyle{acl_natbib}

\newpage
\appendix
\section{Appendix}
\label{app:appendix}

\subsection{Analysis of Activation Token}
\label{analysis_of_activation_token}
\begin{figure*}[!ht]
    \centering    \includegraphics[width=0.8\textwidth, trim=0 0 0 0]{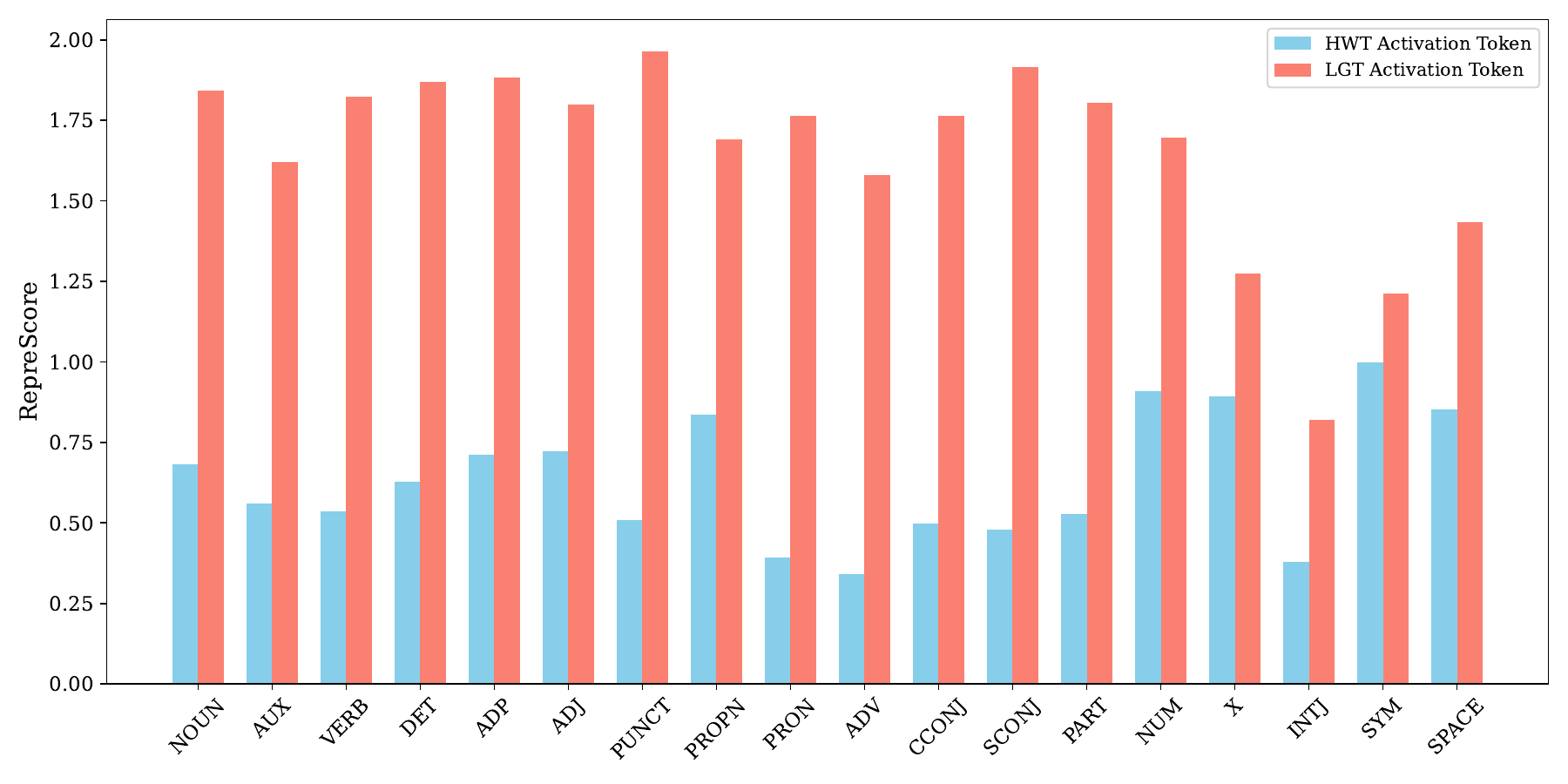}
    \caption{Average RepreScore values across different parts of speech (POS) tags for Activation Tokens. }
    \label{fig:anal_activation_token_pos_vs_score}
\end{figure*}

\begin{figure*}[!ht]
    \centering    \includegraphics[width=0.8\textwidth, trim=0 0 0 0]{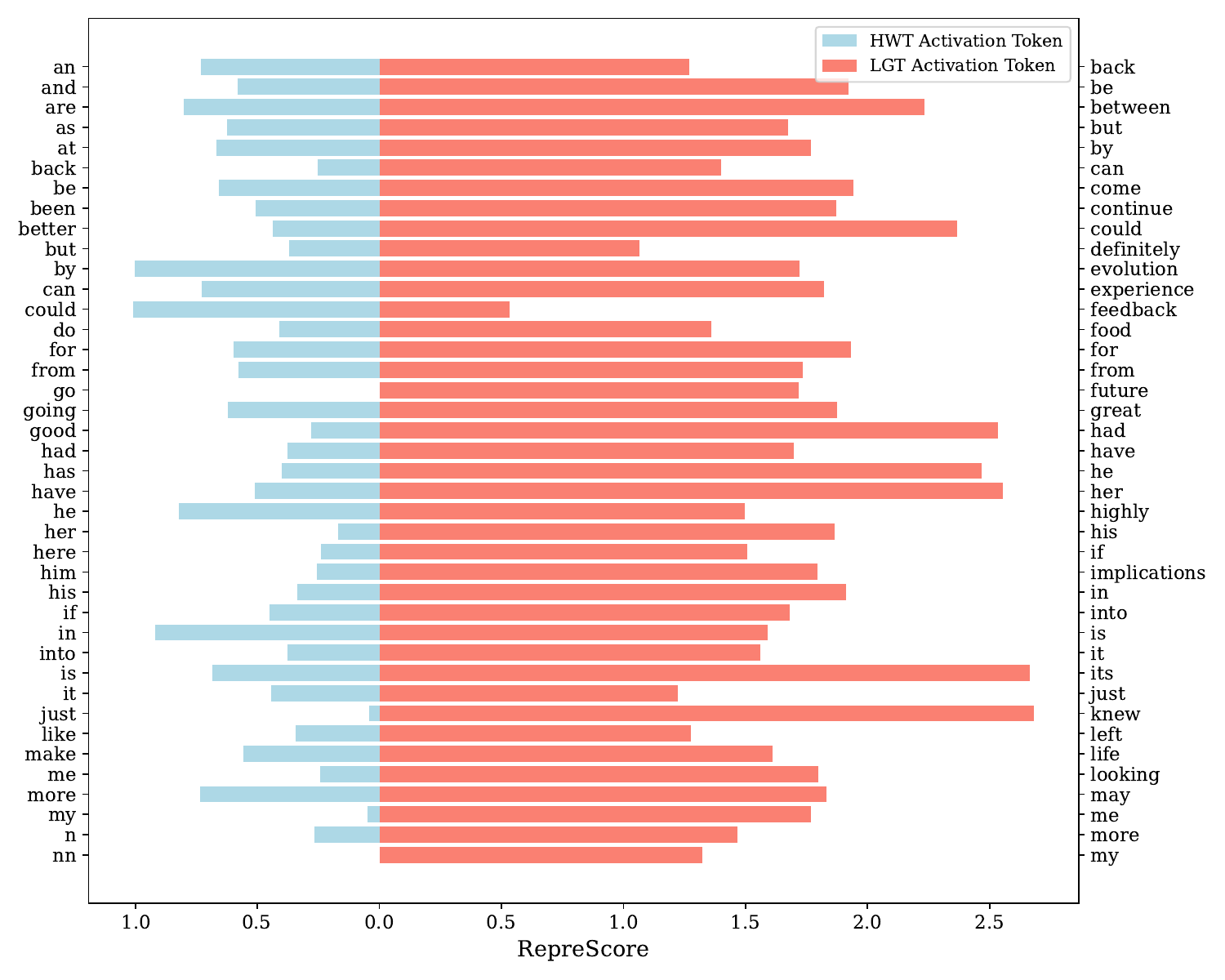}
    \caption{Frequency distribution of the top 50 Activation Tokens and their corresponding RepreScore values for HWT and LGT.}
    \label{fig:anal_activation_token_freq_vs_score}
\end{figure*}

To investigate whether activation tokens contain specific tokens or which parts of speech enable the model to distinguish between HWT and LGT, We conducted an analysis of the word frequency and part-of-speech (POS) tags of Activation Tokens (the last 10\% of tokens) and their relationship with RepreScore. The results are presented in \autoref{fig:anal_activation_token_pos_vs_score} and \autoref{fig:anal_activation_token_freq_vs_score}. In general, the RepreScores for LGT are generally higher than those for HWT, with significant differences observed particularly in adjectives (ADJ), adverbs (ADV) and pronouns (PRON), while the differences for symbols (SYM) are the smallest. From the frequency distribution of the top 50 tokens, it is evident that the same token does not have identical RepreScore values in HWT and LGT; the scores for LGT are generally higher. This suggests that the RepreScore is not directly determined by the token itself. To explain this phenomenon, it is necessary to analyze from the perspective of \autoref{eq:activation}. Since the activation value of each token is computed based on the inputs of its preceding tokens, this implies that when the model processes a token $t_n$, it has already accounted for the contextual information from $\mathcal{T} = \{t_1, t_2, ..., t_n\}$. Therefore, when we calculate the hidden representation changes of a token, these changes are essentially based on an analysis of the complete context rather than an isolated computation of the token itself.

\subsection{Analysis of Model Memorization}
\label{app:analysis_of_model_memorization}
\begin{table}[ht]
\centering
\resizebox{0.25\textwidth}{!}{
\begin{tabular}{l|cc}
\toprule
\bf Train $\downarrow$ & Precision \\
\midrule
ChatGPT & 96.50      \\
Llama-2-70b & 96.50  \\
Google-PaLM & 95.55    \\
Claude-instant & 96.45   \\
Mutil-LLMs & 94.05   \\
AVG. &  95.81   \\
\bottomrule
\end{tabular}
}
\caption{Precision of Different LLMs in Identifying HWT on 2,000 Samples from the Newly Collected Reddit Dataset Released in 2025. Precision is used as the dataset contains only a single class (HWT).}
\label{tab:anal_model_memorization}
\end{table}

Previous research \cite{DBLP:conf/uss/CarliniTWJHLRBS21} has demonstrated the potential to extract substantial portions of text from the training data of LLMs by employing carefully designed prompting techniques. This finding has been further substantiated by subsequent work \cite{DBLP:conf/icml/YuPLDKHLY23}, which introduced advanced strategies for extracting training data. As a result, when text generated by LLMs is sourced directly from their training data, it becomes virtually indistinguishable from human-written text, rendering efforts to differentiate between LGT and HWT content effectively futile. Additionally, recent studies \cite{sun2025aigeneratedtextworldalready} have revealed that a significant portion of contemporary textual data now contains LGT while Reddit has exhibited relatively slower growth in this trend. To ensure that newly collected data consists exclusively of HWT, we utilize the latest Reddit dataset\footnote{\url{https://huggingface.co/datasets/tensorshield/reddit_dataset_157}} released in 2025, which contains content written after the training cut-off dates of the Llama-3.1-8B.\footnote{\url{https://huggingface.co/meta-llama/Llama-3.1-8B}} The results in \autoref{tab:anal_model_memorization} demonstrate that the RepreGuard achieved exceptionally high precision across all model datasets when evaluating new HWT, with an average precision of 95.81\%. This suggests that the RepreGuard is not influenced by model memorization, as the models do not simply recall the texts but can accurately identify the distinguishing features of LGT and HWT texts. 

\subsection{Performance After LGT Pretraining}
\label{app:performance_after_aigc_pretraining}
\begin{figure}[!ht]
    \centering    \includegraphics[width=0.48\textwidth, trim=0 0 0 0]{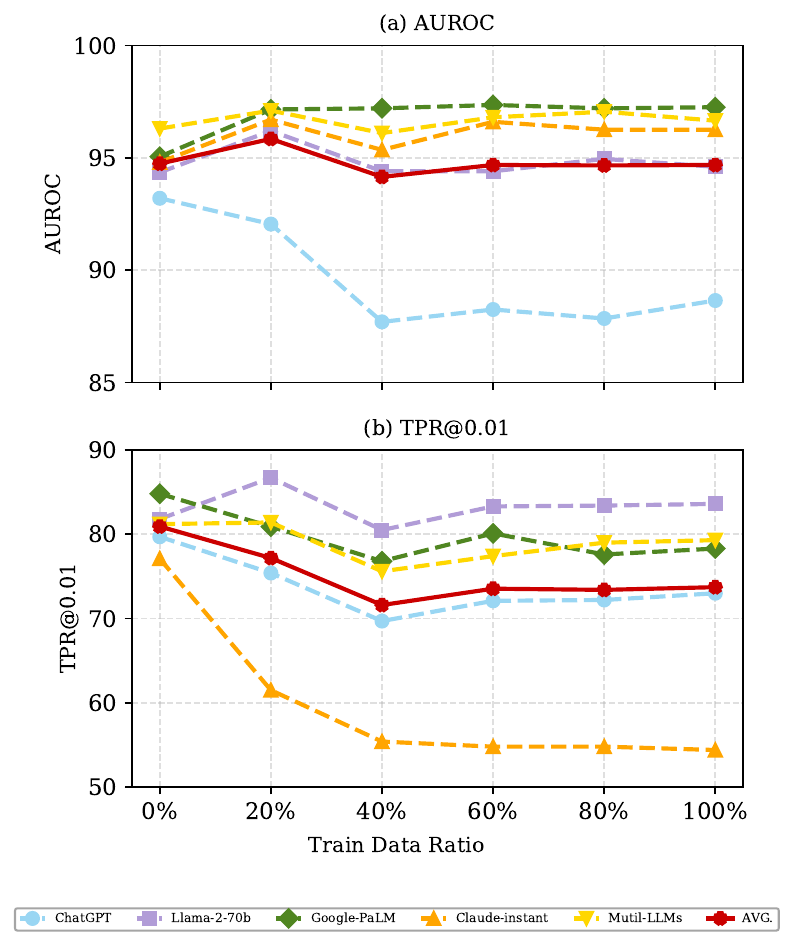}
    \caption{Performance of RepreGuard in Terms of AUROC and TPR@0.01 on 1000 “HWT-LGT” Pairs from 4 Different LLMs after LGT Pretraining. The raw text generated by each model is used to model representation features and set thresholds.}
    \label{fig:performance_after_pretraining_by_LGT}
\end{figure}

As LLMs continue to evolve, an increasing proportion of their training data \cite{DBLP:journals/corr/abs-2306-02707,DBLP:conf/iclr/XuSZG0FTLJ24} is likely to consist of LGT, as the internet becomes increasingly saturated with content produced by these systems. This raises critical questions about the sustained effectiveness of RepreGuard when applied to large-scale datasets in such a scenario. To explore this, we curate a dataset comprising 1 million LGT to pre-train our surrogate model and systematically record the corresponding checkpoints, shown on the figure \autoref{fig:performance_after_pretraining_by_LGT}. The results shown that As the proportion of LGT in pre-training increases from 0\% to 100\%, the AUROC of the RepreGuard exhibits a slight decline, yet overall performance remains robust. Specifically, the average AUROC decreases from 94.76\% to 94.68\%, demonstrating good overall robustness. In contrast, TPR@0.01 experiences a notable reduction, with the average TPR@0.01 decreasing from 80.92\% to 73.72\%. This suggests that pre-training with LGT diminishes the model’s detection capability under extremely low false positive rates. 

\subsection{Discussion on Hallucination Detection and LGT Detection using the Hidden Rrepresentation} 

\begin{figure}[!t]
    \centering    \includegraphics[width=0.50\textwidth, trim=0 0 0 0]{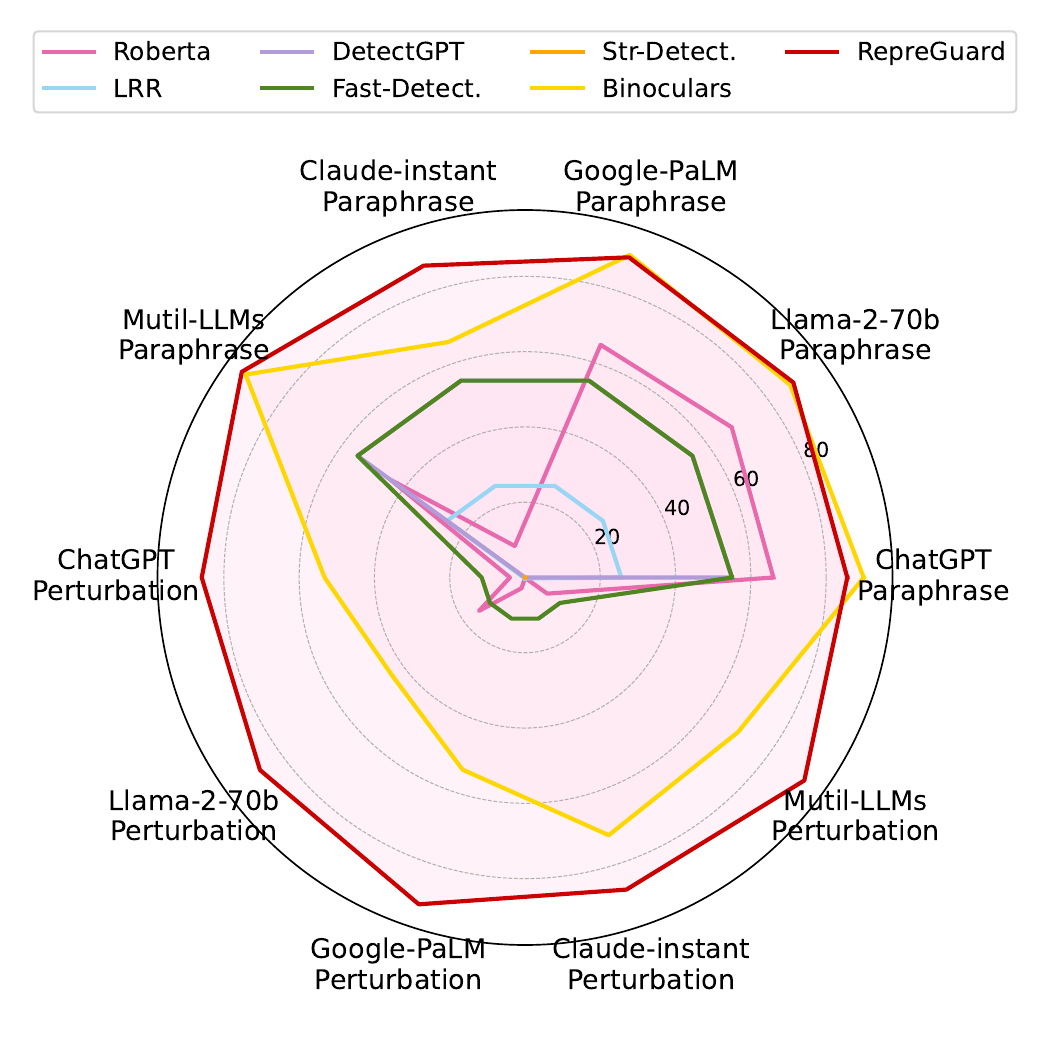}
    \caption{Performance Comparison of RepreGuard on Paraphrase and Perturbation Attack in Terms of TPR@0.01 on 1000 ``HWT-LGT'' Pairs from 4 Different LLMs. The raw text generated by each model is used to model representation features and set thresholds.}
    \label{fig:attacks_tpr}
\end{figure}

Recent hallucination detection research has gradually shifted from focusing on the external performance to the internal 
hidden representation from LLMs. For instance, Masked Grouped Causal Tracing (MGCT)~\cite{monea-etal-2024-glitch} reveals the internal mechanisms underlying grounded and ungrounded behaviors by selectively perturbing and restoring hidden activations. ~\citet{azaria-mitchell-2023-internal} used the hidden representation as feature inputs to train an external feedforward neural network classifier, enabling the automatic determination of statement veracity. The LLM-Check~\cite{NEURIPS2024_3c1e1fdf} method further extracts hidden representations during LLM response generation and calculates the covariance matrix (Hidden Score) as a quantitative metric for hallucination detec tion. These methods collectively validate that the hidden representation contains rich information, offering significant advantages in hallucination detection tasks. 

However, our method, RepreGuard, systematically identifies differences in hidden states between LGT and HWT to distinguish them. Similar to MGCT, RepreGuard focuses on disparities within the hidden space, while MGCT~\cite{monea-etal-2024-glitch} places greater emphasis on causal intervention and mechanistic explanation. In contrast to \citet{azaria-mitchell-2023-internal}, RepreGuard does not require training additional networks, enabling efficient detection within an unsupervised framework. Furthermore, compared to LLM-Check \cite{NEURIPS2024_3c1e1fdf}, which relies on the covariance features of generation, RepreGuard is designed to capture the hidden representations underlying behavioral processes, allowing the model to simulate the writing process and thereby discern differences in hidden representations between HWT and LGT.

\subsection{Figure of TPR@.01 on Paraphrase \& Peturbation Attack}
\label{app:attack_tpr}
~\autoref{fig:attacks_tpr} illustrates the performance comparison of RepreGuard under paraphrase and perturbation attacks in terms of TPR@0.01.

\end{document}